\documentclass[preprint,5p,times,twocolumn]{elsarticle}

\usepackage{amssymb}
\usepackage{amsmath}
\usepackage{xcolor}
\usepackage[bottom]{footmisc}
\usepackage{hyperref}
\usepackage{listings}
\usepackage{graphicx}
\usepackage{booktabs}

\journal{Neurocomputing}

\begin{document}

\begin{frontmatter}

\title{Probabilistic hypergraphs using Multiple Randomly Masked Autoencoders for Semi-supervised Multi-modal Multi-task Learning}
\author[upb,imar]{P\^{i}rvu Mihai-Cristian\corref{cor1}}
\author[upb,imar]{Marius Leordeanu\corref{cor1}}
\affiliation[upb]{organization={Faculty of Automatic Control and Computer Science, Politehnica University of Bucharest}}
\affiliation[imar]{organization={Institute of Mathematics of the Romanian Academy}, postcode={Bucharest}, country={Romania}}
\cortext[cor1]{Corresponding authors. E-mails: \url{mihaicristianpirvu@gmail.com} and \url{leordeanu@gmail.com}}

\begin{abstract}
The computer vision domain has greatly benefited from an abundance of data across many modalities or interpretation layers, in order to improve on various visual tasks. Recently, there has been a lot of focus on self-supervised pretraining methods through Masked Autoencoders (MAE) \cite{he2022masked,bachmann2022multimae}, usually used as a first step before optimizing for a given downstream task, such as classification or regression. The approach proves to be highly efficient and useful in learning as it does not require any manually labeled data. In this work, we introduce Probabilistic hypergraphs using Masked Autoencoders (PHG-MAE): a novel model using multiple modalities or interpretation layers of the input data or tasks, which unifies recent work on multi-layer neural hypergraphs \cite{leordeanu2021semi, pirvu2023multi, marcu2023self, marcu2024quantifying} with the widespread approach of MAEs under a common computational framework. Through random masking of entire modalities, not just patches of those layers, we show that the model samples from the distribution of hyperedges on each forward pass. Additionally, the model adapts the standard MAE algorithm by combining pretraining and fine-tuning into a single training loop. Moreover, our approach enables the creation of inference-time ensembles which, through aggregation, boost the final prediction performance and consistency. Lastly, we show that we can apply knowledge distillation on top of the ensembles with little loss in performance, even with models that have fewer than 1M parameters. While our experiments are focused on outdoor UAV scenes, the same type of tests can be performed in virtually any domain where multiple data layers and modalities are available. In order to streamline the process of integrating external pretrained experts for computer vision multi-modal multi-task learning (MTL) scenarios, we developed an efficient data layer processing and generation software, which we make publicly available. Using this tool, we created and released a fully-automated extension of the Dronescapes dataset, which is the largest of its kind in the literature, to the best of our knowledge. All the technical details, code and reproduction steps of the dataset can be found on the Dronescapes dataset and project's website \footnote{1}.
\end{abstract}


\end{frontmatter}

\footnotetext[1]{\url{https://sites.google.com/view/dronescapes-dataset}}

\section{Introduction}
The world is inherently multi-modal and can be interpreted from multiple points of view. It is desirable and beneficial that all these modalities and interpretation layers work together to find consensus and ensure coherence in the way we understand the world. It makes sense to have a system that combines all these distinct modalities and interpretation layers, in multiple ways, such that we can find consensus among them and obtain a more robust and consistent result. To improve generality, in this work we use interchangeably the terms \textit{modality} (e.g. visual, sound or language) and \textit{interpretation layer} (e.g. semantic segmentation), as they are both referring to different ways to measure and perceive the world. The term "task" will specifically refer to the output modality or interpretation layer, not given at test time.

Previous works that perform self-supervised multi-modal multi-task learning model the structure of the data as a multi-layer graph or hypergraph \cite{leordeanu2021semi,zamir2018taskonomy,marcu2023self,pirvu2023multi,haller2021self}. There are two key issues with these approaches: \textbf{1)} they have a fixed hypergraph structure, which must be defined beforehand by hand and \textbf{2)} each edge or hyperedge has a dedicated unique neural network. Due to this, training the whole graph or changing its structure becomes highly expensive and laborious.

Our proposed approach mitigates both these limitations by having a single multi-layer neural network that models the whole hypergraph with a masked autoencoder approach, which samples from the space of hyperedges and forms hyperedge ensembles. As we will show later, each forward pass through this network is in fact a pass through a corresponding hyperedge enabled by a specific random masking.

Masked Autoencoders (MAE) were initially introduced in Natural Language Processing \cite{devlin2019bert} and later ported with great success to computer vision \cite{he2022masked}. Today they are mostly used during a self-supervised pretraining stage, which must be followed by task-specific fine-tuning or linear probing. In this work, in order to extend the standard MAE approach to our case of multi-modal multi-task learning, we define two disjoint sets of "input" and "output" modalities, such that the input modalities could be available (all or in part) during test time, while the output ones are not. To better distinguish between the two types, we will also refer to the output modalities as "tasks".

Conceptually the two sets of input and output modalities create a bipartite graph. For example, low-level modalities (i.e. RGB, HSV) are only inputs, while high-level ones (depth estimation, semantic segmentation) are always predicted during test time. In this way the model is trained to learn only relevant dependencies between modalities (i.e. inputs predicting outputs, not the other way around). By predefining a set of input and output modalities, as well as task-specific losses (e.g. cross entropy for classification and MSE for regression), we successfully combine pretraining and fine-tuning into a single training loop, from scratch, without the model getting stuck in local minima, as it often happens with a standard MAE, while simplifying the process. Furthermore, we mask the entire modality, not just at patch level, which helps with performance metrics.

We call our method Probabilistic hypergraphs using Masked Autoencoders (PHG-MAE) and it aims to bridge the classical Neural Graph methods with the more modern Masked Autoencoders. In order to do this, we extend the proxy task of Masked Autoencoding (MAE) where each random masking is equivalent to sampling one specific hyperedge at a time. By chaining multiple inference steps together, our model converges to sampling from the full graph defined by all the combinations of inputs and outputs. The model is described in greater detail in Section \ref{sec:phg-mae}.

In machine learning, tasks are aligned if their underlying prior distributions are related (e.g., by having a small KL divergence). If so, learning a transformation between them requires fewer data samples and less model complexity. For example, predicting camera normals from RGB is less aligned than from depth sensors. The first needs complex networks and large datasets, while the second can approximate analytical solutions, potentially using smaller networks and fewer samples. In multi-modal systems, tasks may have varying alignment and training only a standard MAE often leads to uneven performance. This is also related to the concept of negative transfer in MTL \cite{wu2020understanding} where competing tasks may hurt each other's performance. To mitigate these complexity differences between tasks, we introduce derived intermediate modalities. These act as a bridge between low level inputs (like RGB sensors) and high level outputs (like segmentation or depth), simulating curriculum learning. This, alongside the probabilistic hypergraph approach and ensembles, allows the model to learn optimal input-output dependencies from data without explicit manual modeling.

While our aim is to learn a model that solves multiple "tasks" (multiple output layers), we focus a large portion of this work on semantic segmentation, being a high-level task of widespread interest, impacting many AI fields.
However, semantic segmentation, as it is currently tackled, has some inherent issues. For example, the list of classes is fixed and, more than that, they can be at different semantic generality levels. For example, we could have as classes both 'residential area' as well as 'chair', where one is a generic concept, while the latter is a concrete object. This dichotomy has been previously studied with solutions such as separating classes in "stuff" (more general) and "things" (concrete objects) \cite{lin2014microsoft}. Moreover, the same class can be interpreted depending on context. For example, a 'building' can be interpreted from an architectural point of view, as opposed to 'houses', while it can also be interpreted as 'school'. One observation that we make in this work is the fact that the generic classes (i.e. residential, water, sky, vehicle) are more robust to domain shifts compared to more specific ones (car model, oil rig, mountain bike etc.). For this reason, we export various general modalities such that our model can have access to these intermediate representations as a bridge between low-level raw signals (RGB) and high-level semantics.

One theme of this research was the ability to quickly iterate and add new modalities derived from pretrained experts. We believe that only through high quality and diverse data we can achieve training efficiency as well as strong and robust ensembles. To achieve this, we developed a data-pipeline that allows us to quickly add or remove experts as well as procedural derivatives from experts for any arbitrary video (in the domain of UAVs, autonomous driving or similar robotics-related domains). We use the original Dronescapes videos as a starting point, but also include new UAV videos to test the hypothesis that through new data augmented with experts we can improve the quality of the predictions on unseen test scenes. We release the code of the data-pipeline as an open-source project.

\textbf{Focus on multi-modal real-time AI for the real world:} As the AI field enjoys an immense flourishing period with applications in virtually all domains of the society, there is still a great need for real-time low-cost solutions that learn and function robustly and efficiently in the multi-modal world. Our approach is focusing on this important practical aspect, for which we provide an algorithmic solution with demonstrated experimental results as well as a fast and low memory cost implementation, for layer generation, training and testing, which we make available on the project website.

\textbf{The main contributions} of this work are:
\begin{enumerate}
	\item Probabilistic hypergraphs using Masked Autoencoders (PHG-MAE): an extension of the standard MAE pretraining algorithm that enables Multiple Randomly Masked Autoencoder Ensembles at inference time and unifies previous hypergraph methods under a single neural model.
	\item Merging of pretraining and task-specific fine-tuning under a single training loop by defining inputs and outputs and influencing the masking algorithm accordingly. Furthermore, we only mask an entire modality, not at patch level as previously done.
	\item Inclusion of procedurally derived intermediate modalities from pretrained experts on diverse datasets to leverage their knowledge and smooth the learning difficulty from low-level inputs to complex high-level tasks as they are more general and robust to domain shifts across scenes.
    \item Efficient training and distillation showcasing competitive performance and state-of-the-art video consistency with small and very small (150k $\sim$ 4.4M) CNN networks on both the Dronescapes multi-modal multi-task benchmark and unseen videos, enabling research on commodity hardware. For this, we introduce a fully automatic extension (30K $\rightarrow$ 150K samples) of the original Dronescapes dataset with a path towards massive dataset generation from raw videos only.
	\item An open-source data-pipeline that enables efficient extraction of new modalities from pretrained experts on videos. This simplifies and automates the process of creating large video datasets for training semi-supervised multi-task vision models for aerial image understanding and beyond.
\end{enumerate}

\section{Related work}

We provide a short summary of related work and state-of-the-art in a few key areas that our research touches upon for context.

\textbf{Multi-modal multi-task datasets:} data is the fuel that powers any machine learning system. For this reason, having good, reliable and curated datasets, annotated for one or multiple tasks is a requirement for building such systems and doing research on models and applications. The release of strong and unbiased benchmarks has driven the progress in the domain. This subsection provides an overview of the landscape of datasets for multi-modal multi-task learning (MTL) in computer vision and robotics. In the space of autonomous driving with real footage there are datasets such as KITTI \cite{geiger2013vision}, the Cityscapes dataset \cite{cordts2016cityscapes}, Open MARS \cite{li2024multiagent}. Synthetic datasets include AIODrive \cite{weng2021all}, RealDriveSim \cite{jadon2025realdrivesim}, SHIFT \cite{sun2022shift} using the CARLA simulator \cite{dosovitskiy2017carla}, Zenseact Open Dataset \cite{alibeigi2023zenseact} or WaterScenes \cite{yao2024waterscenes} which facilitates autonomous driving on water. Moving on to indoor robotics and scene understanding, there are datasets such as NYUDepth \cite{eigen2014depth}, THUD++ \cite{li2024thud++}, REASSEMBLE \cite{sliwowski2025reassemble}, a place and anomaly detection dataset \cite{wozniak2025multi}, RoboCasa \cite{nasiriany2024robocasa}, BridgeData V2 \cite{walke2023bridgedata}, Lambda \cite{jaafar2024lambda} or RoboMNIST \cite{behzad2025robomnist}. In Earth observation there are multi-modal datasets such as NEO \cite{pirvu2023multi}, M3LEO \cite{allen2024m3leo}, TerraMesh \cite{blumenstiel2025terramesh}, MDAS \cite{hu2023mdas} or Ticino \cite{barbato2024ticino}. Continuing to aerial image understanding and UAVs, there are datasets such as Dronescapes \cite{marcu2023self}, UEMM-Air \cite{yao2024uemm}, UAV3D \cite{sunderraman2024uav3d}, MMAUD \cite{yuan2024mmaud}, MMFW-UAV \cite{liu2025mmfw}, SynDrone \cite{rizzoli2023syndrone}, DDOS \cite{kolbeinsson2024ddos}, AU-AIR \cite{bozcan2020air}, HDIN \cite{chang2022hdin}, AirV2X \cite{gao2025airv2x} or University-1652 \cite{zheng2020university}. Recently, there has been a survey covering the topic of synthetic vs. real datasets \cite{marcu2024quantifying} targeted at this domain. They argue that the gap between real datasets and synthetic ones is still too large, introducing a new metric for perceptual and structural disparities between domains. This makes offline training followed by generalization to real data still challenging and they argue that the simulators must be further improved in realism.

Most of the works presented in this section either require human annotation, work on simulated data or target tasks where two or more sensors (such as location and images) are captured at the same time. In this work, we extend the Dronescapes dataset \cite{marcu2023self} with additional modalities and new videos, resulting in an increase from $\sim$23K samples to a total of ~148K automatically annotated samples ready to use for training MTL models. We extend the dataset with various new modalities from pretrained experts: semantic segmentation, depth estimation, optical flow, and derived ones, including camera normals, different binary segmentations, safe landing areas etc. We hope that our data-pipeline can enable greater automation of semantic and structural labeling for videos in the wild, not only in the UAV domain but also for other settings, such as indoor robotics videos or autonomous driving footage.

To our best knowledge, the extended Dronescapes dataset introduced here, which we made public,
becomes the largest multi-modality video dataset for UAV vision research.

\textbf{Multi-modal multi-task learning (MTL):} multi-modality refers to handling multiple input signals, while multi-task is defined by a single model predicting two or more tasks at once (i.e. semantic segmentation and depth estimation). Recently, there has been a surge in large models with multi-modal abilities, however they are usually limited to just two modalities: text and RGB images \cite{zhang2024vision}. For multi-task prediction, \cite{standley2020tasks} proposes a multi-encoder architecture where semantically similar modalities share a network branch, while competing ones are assigned to separate branches to mitigate negative transfer. The work of \cite{wu2020understanding} also explores the negative transfer phenomenon, while \cite{liu2019loss} propose a mitigation by dynamically weighting up or down each task during the training based on the running error. In \cite{leordeanu2021semi}, they introduce a graph-based MTL system, where each graph edge represents a neural network modeling two modalities (one input and one output), forming ensembles through multiple pathways to the same destination task. Later \cite{marcu2023self} extends this paradigm to real world data and hypergraphs. The work of \cite{zamir2018taskonomy} presents an automated framework for discovering semantic taxonomies based on related modalities. MTL has also been applied to specialized fields, including Earth observation \cite{pirvu2023multi}, medical imaging \cite{haque2021multimix} or recommender systems for videos \cite{li2023adatt}.

In this work we employ a MTL setup for our experiments based on the tasks of the Dronescapes dataset: semantic segmentation, depth and camera normals estimation. We also define a fixed set of input and output modalities aligned with the original Dronescapes dataset, always keeping RGB as input. To our best knowledge, our model has the largest number of modalities (inputs, intermediate and outputs) in the multi-modal vision literature.

\textbf{Graphs and hypergraphs in machine learning:} graphs have been a longstanding area of research from mathematics to computer science and now to machine learning. Graphs in machine learning have been popularized since the introduction of Probabilistic Graphical Models \cite{koller2009probabilistic,bishop2006pattern}. In this framework, complex data distributions are defined through the view of the Bayesian principles of cause (prior) and effect (posterior) and Bayesian Networks. Then, in the Neural Graph Consensus (NGC) model \cite{leordeanu2021semi}, graphs are constructed on top of neural networks, where an edge explicitly models the relationship between two modalities in a MTL system. Later, this work is extended to hypergraphs in \cite{marcu2023self,pirvu2023multi}, where hyperedges model the relationship of sets of nodes. Graphs are used to model other types of problems as well. The work of \cite{kipf2016semi}, introduces the semantics of Graph Neural Networks as an extension of the Convolutional operator on unstructured grids, which is then applied to specialized domains such as social networks \cite{shlomi2020graph}, modeling particle physics systems, in a field usually called Geometric Deep Learning \cite{bronstein2021geometric}. The combination of MTL and graphs in machine learning enables a holistic modeling and understanding of the data. For example, in Earth observation many layers from satellites can be combined together to make inferences about future states of our planet, like weather, fires, etc. \cite{reed2023scale,nedungadi2024mmearth,guo2024skysense}.

In this work we extend the work on multi-modal (multi-task) neural hypergraphs by improving on one of its core weaknesses: the rigid structure of the graph which must be predefined. We employ a single neural network which, through the lens of probabilistic sampling using Masked Autoencoders, models the entire hypergraph of relations between pairs of nodes in a MTL system. Our model is thus suitable for datasets where a holistic understanding between multiple modalities is desired.

\textbf{Masked Autoencoders (MAE):} MAE was initially introduced in Natural Language Processing \cite{devlin2019bert} and later ported with great success to computer vision \cite{he2022masked}. Recent works, such as \cite{bachmann2022multimae}, leverage MAE-based solutions for learning depth estimation and semantic segmentation via self-supervised pretraining followed by task-specific fine-tuning. \cite{lu2024unified} and \cite{mizrahi20234m} extend this approach to new modalities, including image, text, audio generation, and action prediction for robotics. These advances are driven by the rise of foundation models pretrained on massive datasets, enabling zero-shot prediction via prompting and efficient fine-tuning, as seen in \cite{radford2021learning} and \cite{kirillov2023segment}. \cite{bai2022ofasys} proposes a multimodal, promptable model with an instruction-based domain-specific language for generalist capabilities across text, images, audio, and video.

In this work, we design a new MAE-based algorithm for MTL, called Probabilistic hypergraphs using Masked Autoencoders (PHG-MAE), which we describe in greater detail in the next section. The main extension of the classic MAE is the random maskings of entire input modalities which enable the formation of robust ensembles, which also provide reliable pseudolabels for the self-supervised learning stage.

\textbf{Ensemble learning and inference-time ensembles:} ensemble learning \cite{ganaie2022ensemble,mienye2022survey} is a longstanding research topic in machine learning that helps boost both prediction performance and consistency. It is a method that tackles the problem of prediction consensus and model uncertainty from different predictors \cite{gawlikowski2023survey}. It consists of having multiple models that independently predict the same outputs for a given task. Then each of these independent predictions is combined into a final prediction using an aggregation function. It turns out that having different models is not the only way to get ensemble candidate predictions. For example, if a model has some random process in its algorithm, then we can get multiple predictions for the same input if we do multiple inference-time passes. This can be achieved by either having softmax with temperature at the output layer, having dropout enabled at inference time \cite{inoue2019multi, gal2016dropout} or having parts of the inputs randomly modified (i.e. augmentation): random augmentations \cite{kim2020learning, shanmugam2021better}.

As mentioned above, we also perform random input masking combined with Masked Autoencoders. We mask whole input modalities, not just at patch level, as in previous works, a strategy that mimics the learning of various hyperedges. Then, each reconstruction is added to the pool of ensemble candidates and this process is repeated many times until we achieve consensus across predictions. We describe this algorithm in more detail in Section \ref{subsec:random-masking-ensembles}.

\textbf{Semi-supervised learning and model distillation:} semi-supervised learning is a setup where both labeled and unlabeled data are available. As human-annotated data is hard to produce, and algorithmic labels are not available or reliable for most tasks, this technique is of utmost importance. In this case, a separate process (e.g., one or more pretrained models) is applied to the unlabeled set to generate pseudo-labels. Then, in a secondary step, these pseudo-labels, alongside the labeled data, are then used to train a new model with the idea that the additional pseudo-labels may provide extra signal to boost performance compared to a purely supervised setup. This approach has been applied successfully in object recognition \cite{shehzadi2024semi}, as well as in open environments with large class imbalance \cite{gui2024survey,guo2025robust}. Moreover, when a supervised model is used to produce pseudo-labels (even on the labeled dataset) this is known as teacher-student model distillation \cite{hinton2015distilling}. In this setup, the teacher model (usually a larger one) captures the data distribution, while the student model is trained to mimic the teacher’s learned distribution. This method is often used to build smaller models for real-world deployment, where the larger model is only used to generate pseudo-labels. Combining semi-supervised learning with teacher-student distillation has been a very successful technique, with great results on image classification \cite{laine2016temporal,tarvainen2017mean,berthelot2019mixmatch} as well as foreground object segmentation \cite{croitoru2019unsupervised} with little to no labels available.

In this work we perform semi-supervised learning at various stages. First, we use pretrained experts and our data-pipeline to generate pseudo-labels on new and unseen videos similar to the ones in the original Dronescapes dataset for all three tasks (semantic segmentation, depth estimation and camera normals estimation) as well as various intermediate modalities (i.e. buildings, sky, water etc.). Using these initial expert and derived pseudo-labels, we train our PHG-MAE model by creating a multi-modal multi-task learning (MTL) model. Then, in a subsequent step, we do iterative semi-supervised learning through model distillation by generating pseudo-labels for semantic segmentation using our MTL model on yet another set of new and unseen videos. This step goes in the direction of creating small and very small RGB input-only models (150k and 430k parameters) with little degradation in performance, useful for real-time semantic segmentation on commodity hardware.

\textbf{Test-Time Adaptation (TTA):} there is a recent line of research which argues that the current state of machine learning where we have two steps: training on large amounts of data followed by deployment and inference using frozen weights and no further learning is a big bottleneck towards major breakthroughs in AI. This is especially true as the data distribution changes in the real world compared to the fixed dataset the model is trained on. TTA tries to mitigate this by allowing the model to adapt for each test sample \cite{liang2025comprehensive,xiao2024beyond}. There are benchmarks designed specifically for this kind of adaptation to novel tasks \cite{chollet2025arc} showcasing that models with only 'memorization' layers (i.e. standard supervised learning) fail to adapt. There are various strategies for TTA, such as test-time augmentation \cite{wang2024comprehensive} or pseudo-labels generated from the test samples followed by test-time learning \cite{karimi2025adapac}. Learning at test time can also be expensive and destructive (i.e. catastrophic forgetting), needing to either remove specialized layers adapted to the original train set, like batch normalization statistics \cite{su2024towards}. Recent models, especially in the NLP domain are 'promptable' without any re-training or fine-tuning required, enabling zero-shot performance on novel tasks \cite{radford2019language}. These promptable models can be queried multiple times followed by an aggregation step, which results in an ensemble-like method. These are also called 'reasoning models' \cite{snell2024scaling,beeching2024scalingtesttimecompute}.

In this work, we design a form of TTA, by doing multiple independent queries through random maskings of vision modalities followed by an aggregation. We also explore a bit the area of guided search through finding a more optimal masking distribution compared to the default uniform distribution across modalities.

\section{PHG-MAE: Probabilistic Hypergraphs using Masked Autoencoders}
\label{sec:phg-mae}

In this section we present in technical detail the new approach which unifies, as mentioned in the introduction, the different computational concepts masked autoencoders, multi-modal hypergraphs, ensembles and semi-supervised multi-task learning. These are the basic components that, when put together, define our PHG-MAE model.

\subsection{Multi-modal Neural Graphs and Hypergraphs}
\label{subsec:neural-graphs}

\begin{figure*}[h]
    \centering
    \includegraphics[width=0.8\linewidth]{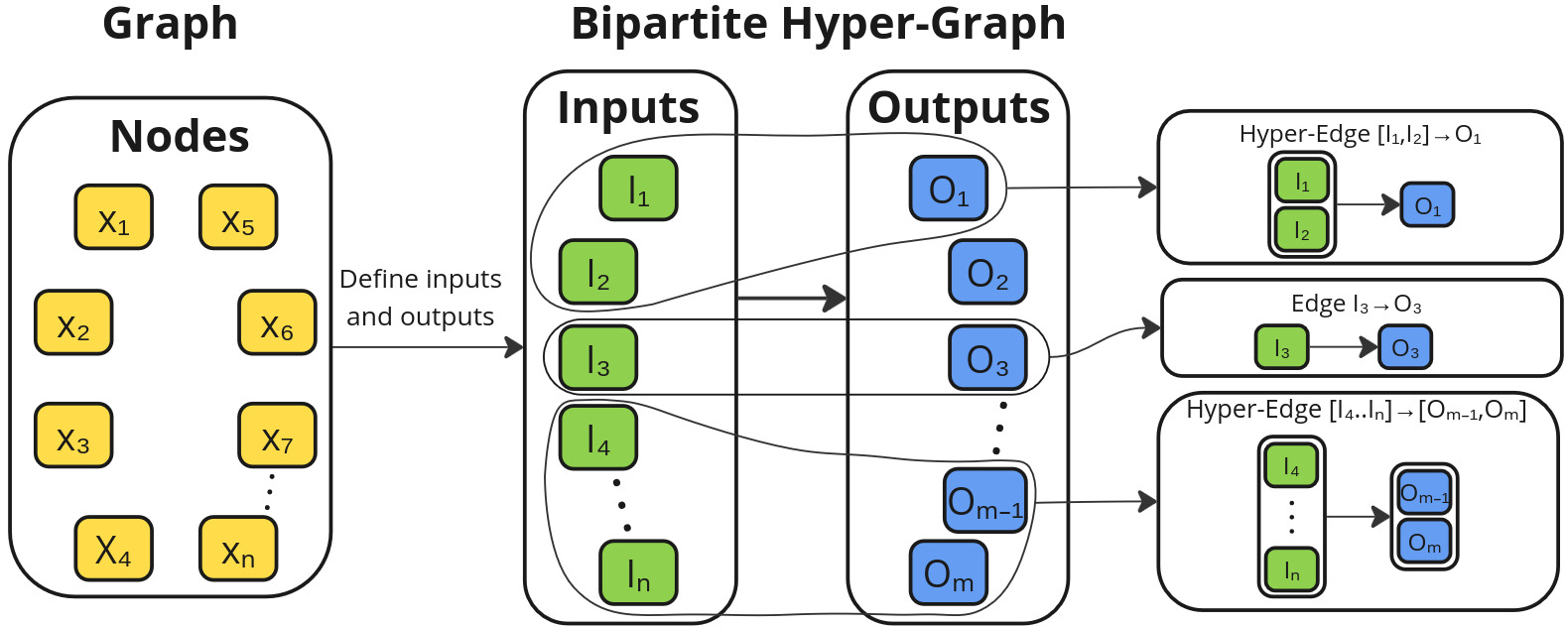}
    \caption{From a graph to a bipartite hypergraph: defining inputs and outputs, followed by creating hyperedges from many input nodes to one or more output nodes. Three edges (out of many valid ones): $[I_{1},I_{2}] \rightarrow O_1$ (hyperedge, top right), $I_3 \rightarrow O_3$ (edge, middle right) and $[I_4,..,I_n] \rightarrow [O_{m-1}, O_m]$ (multi-output hyperedge, bottom right)}
    \label{fig:hypergraph-basic}
\end{figure*}

In Figure \ref{fig:hypergraph-basic} we present the basic idea of a neural hypergraph.

This notion of neural hypergraphs was first introduced by NGC \cite{leordeanu2021semi} and then extended in \cite{marcu2023self,pirvu2023multi}. A graph is mathematically defined as $G:(V, E)$ with $V$ being the set of vertices or nodes and $E$ the set of edges. A bipartite graph groups the nodes in two disjoint sets: input nodes $I:(I_1, I_2, ..., I_n)$ and output nodes $O:(O_1, O_2, ..., O_m)$. Traditionally, these nodes represent modalities, such as RGB, Depth estimation, Semantic segmentation etc. However, the concept of graphs is not limited to a particular application, and they can represent anything as long as it models the problem at hand (i.e. patches of an image, words in a sentence, user profiles in social media etc.). A neural graph is defined by a graph whose edges are neural networks, meaning that the transfer function between two nodes $Edge:(I \rightarrow O, f_{nn})$, where $I$ is an input node, $O$ is an output node and $f_{nn}$ is a non-linear multi-layer transformation, different from the work of GNNs \cite{kipf2016semi}, where the function is a set of non-linear shallow transformations applied multiple times. A hypergraph is a graph where one hyperedge contains more than one node on either side of the transformation: $Hyperedge:(\{I\} \rightarrow \{O\}, f_{nn})$, where $\{I\}$ is a set of one or more input nodes.

Then, we introduce the concept of a pathway from one input node to one output node which is a succession of edges (neural or not). Single-hop edges, represent a pathway of length one, for example Edge $I_1 \rightarrow O_1$. Similarly, we can have single-hop hyperedges, such as $[I_1, I_2] \rightarrow O_1$ where we first concatenate the two input nodes together followed by a transformation towards the output node. These are called \textit{Aggregation Hyperedges} (AH) in the original work. And, of course, we have a single-hop hyperedge from one input node towards multiple output nodes $I_1 \rightarrow [O_1, O_2]$. In the original work these edges are represented by independent deep neural networks, however this is not always needed. As long as we can model the transformation, any function can be applied. Furthermore, there can be pathways of length greater than one, such as two-hop edges: $I_1 \rightarrow O_1 \rightarrow O_2$ where we have two transfer functions (i.e. neural networks), where the second one depends on the first one to be computed or learned.

The main limitation with the graphs and hypergraphs methods for MTL introduced in other works is the need to predefine the structure of the graph beforehand. This means that all the edges, such as the three showcased in Figure \ref{fig:hypergraph-basic} right side, must be fixed. Once settled, these can no longer be changed, as the ensembles used in their methods depend on the data distribution of those exact edges in that exact order. Any change at hypergraph structure level requires significant downstream retraining. Moreover, the hypergraph can grow very large. When trained to predict earth layers for the NEO dataset \cite{pirvu2023multi}, they used up to 126 individual neural networks, which required designing an entire framework for proper management, adding complexity. Motivated by these bottlenecks, we formulate the hypergraph through the lens of probability distributions using a single neural network. Moreover, in this work we focus only on pathways of length one.

\subsection{Modeling Multi-modal Neural Hypergraphs with Masked Autoencoders}
\label{subsec:phg-modeling-hg-with-mae}

\begin{figure*}[h]
    \centering
    \includegraphics[width=1\linewidth]{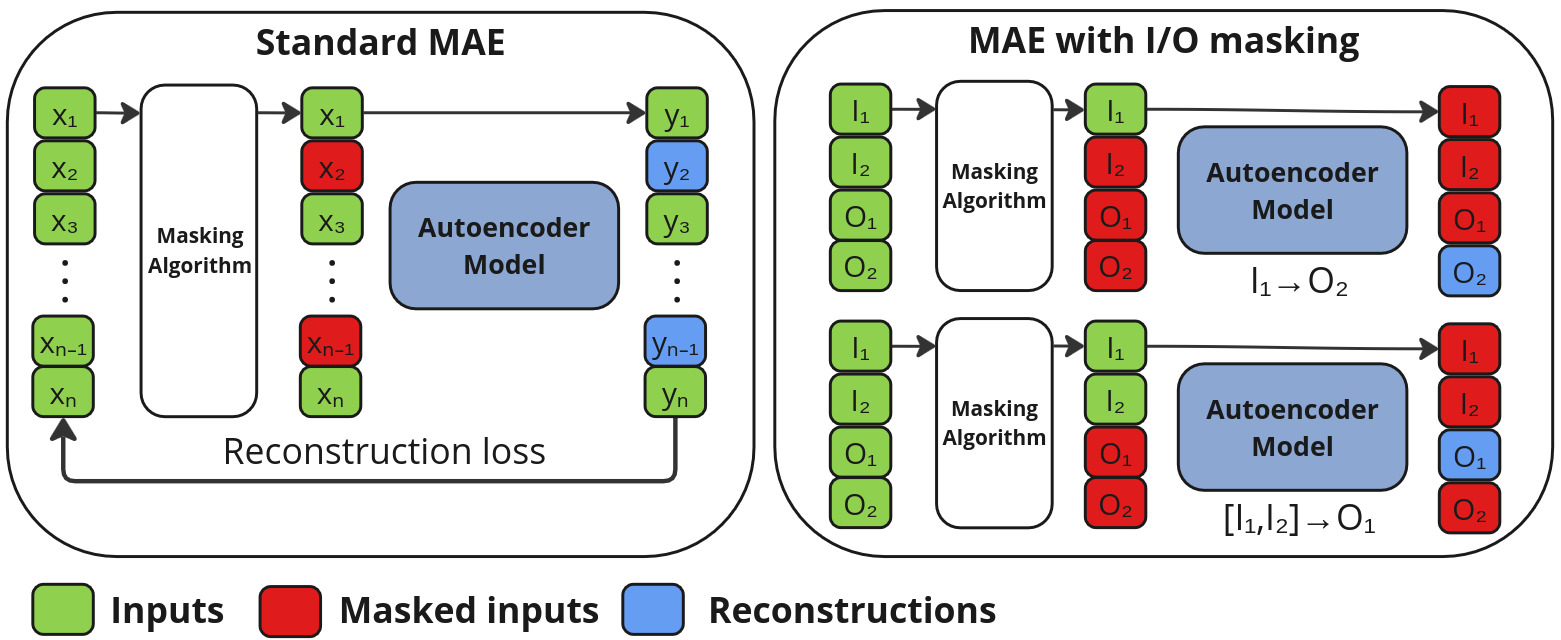}
    \caption{Left: Standard Masked Autoencoder model. Right: Masked Autoencoder model with explicit masking rules that model certain input and output pairs which teach the model about the distribution of hyperedges. Two such masking operations are shown (out of many valid ones): $I_1 \rightarrow O_2$ (top right), and $[I_{1},I_{2}] \rightarrow O_1$ (bottom right). Inputs are either seen or masked, but never reconstructed, while outputs are always masked and sometimes reconstructed.}
    \label{fig:mae-vs-io-mae}
\end{figure*}

In Figure \ref{fig:mae-vs-io-mae}, we present our key modification to the Masked Autoencoder model.

We start with a regular Masked Autoencoder model (left side). At this point, we don't have any input or output modalities defined, so we treat all of them as generic input features $X:(x_1, x_2, ..., x_n)$. These input features can be patches of an image like in ViT \cite{dosovitskiy2020image} or even words in sentences like in the original BERT MAE \cite{devlin2019bert}. We can already observe some similarities compared with Figure \ref{fig:hypergraph-basic}, however, the nodes are now stacked together to fit in the MAE model, and, with proper masking, we can emulate all the possible hyperedges of the original graph. In our case, we mask entire modalities, each corresponding to a full image. The outputs of the model are also vector features $Y:(y_1,y_2, ..., y_n)$ with one caveat: only the masked ones (red) are reconstructed, while the unmasked ones are simply copied as is. This forces the network to only learn reconstructions (not data copying), which boosts the final performance, aligned with the results also observed by \cite{he2022masked}.

On the right side of the figure, we present two samples based on the modifications to the masking algorithm which allows combining pretraining and fine-tuning in a single training loop. For this purpose, we propose some key modifications: 1) creating a custom masking algorithm $M$ that uses the notion of input and output nodes and 2) defining task-specific loss functions for each output node.

\textbf{\underline{Proposition}} A forward pass through PHG-MAE is equivalent to a forward pass of a single hyperedge in a hypergraph. Formally, this can be defined as: $HE \sim M(PHG\text{-}MAE)$, where $HE$ is a sampled hyperedge from the hypergraph $PHG\text{-}MAE$ with the masking algorithm $M$. By doing multiple maskings using the $M$ function, we eventually recover all the possible hyperedges, thus recreating the entire graph. This is possible as long as the masking algorithm is implemented such that the desired graph structure is respected.

\textbf{Define inputs and outputs} Similarly to the previous works on graphs (NGC) and hypergraphs (NHG), we define two disjoint sets of inputs and outputs. As a rule of thumb, if a dataset doesn't explicitly enforce them (i.e. real world constraints), we set as inputs those modalities that are easy to acquire (i.e. RGB) and as outputs those that are hard to acquire (i.e. semantic segmentation which may require human annotation). To achieve this with a MAE model, the output modalities are \textit{always masked}, while input modalities are \textit{always seen}. We always mask at whole modality level, not at patch level, though this is not a necessity depending on the task at hand. This marries the standard input-output prediction paradigm where, for example, RGB is always used as input and the semantic segmentation is always predicted with the auto-encoder paradigm, where we would usually both encode and reconstruct all the input features.

It should be noted that the disjoint input and output distinction is not a necessity, we can very well train a model using the standard MAE approach where any features can reconstruct all the other ones. However, when doing this, the model also learns "impossible" (or impractical) combinations, like combining input and output modalities together instead of one reconstructing the other. As the output modalities are usually more high level (i.e. semantic segmentation or depth estimation), the models learn to rely on them more than on the low level cues (like RGB images or simple edges), getting stuck in local minima and underperforming in test scenarios, where only input modalities are available. If we train a model in this standard MAE regime, it requires fine-tuning as a secondary step where inputs and outputs are distinguished. Therefore, while this separation is not technically needed, it is a key component that allows us to do a single training loop without any secondary task-specific fine-tuning.

\textbf{Task-specific loss function:} we perform more than just regression-based reconstruction as it is usually done in standard MAE. Instead, we treat each task independently and provide task-level loss functions. For example, for the semantic segmentation task, we use the cross entropy loss, while for other regression-based tasks, such as depth or camera normals prediction, we use the standard L2 loss. Note that for depth estimation we could go even further and include photometric loss \cite{zhou2017unsupervised} while for camera normals, we could also derive them from the estimated depth using algorithms based on SVD \cite{hartley2003multiple}. Generally speaking, each task can have its own set of loss functions best aligned with it. In this work, we only use the L2 loss mostly due to its generality and simplicity, but we acknowledge that there may be better suited ones.

By combining these two key modifications and through multiple random maskings, we are sampling from the distribution of all hyperedges, thus modeling the original neural hypergraph with a single Masked Autoencoder model.

\subsection{Multiple Randomly Masked Autoencoder Ensembles}
\label{subsec:random-masking-ensembles}

\begin{figure*}[h]
    \centering
    \includegraphics[width=1\linewidth]{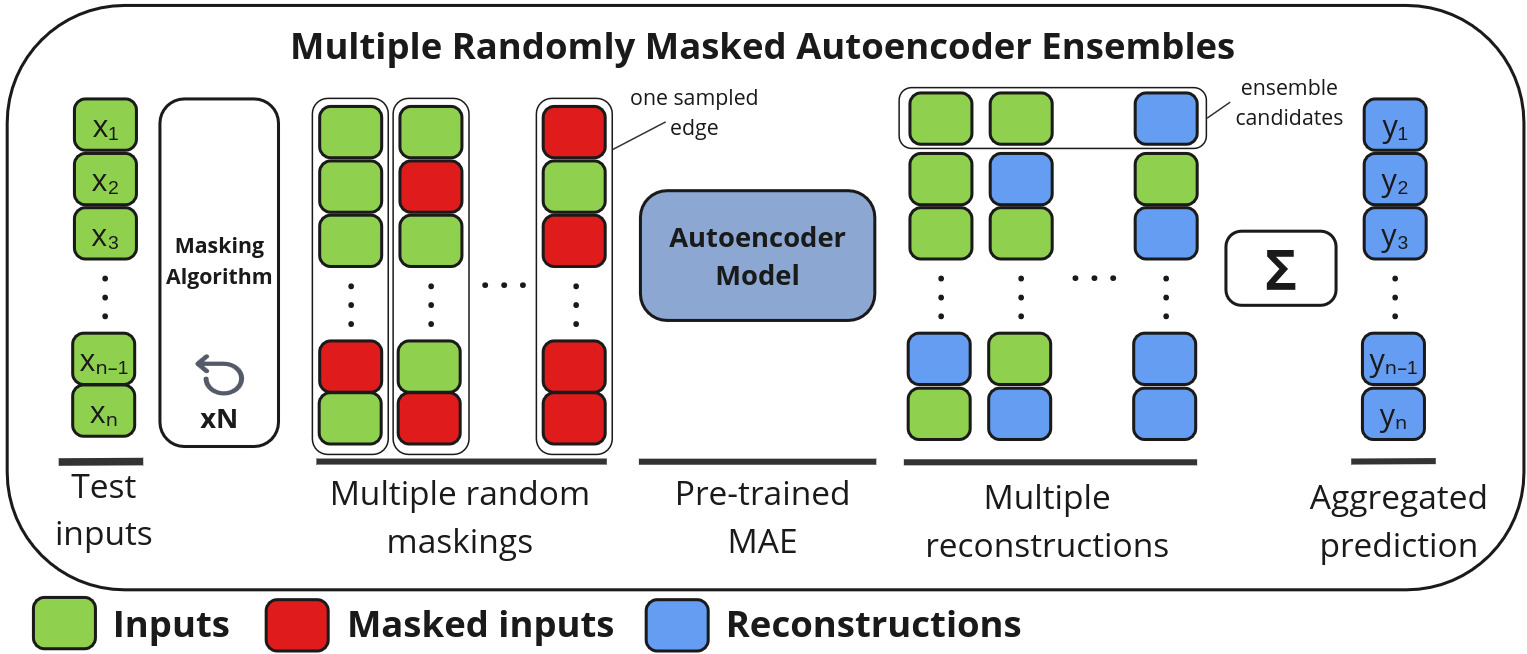}
    \caption{We use a pretrained MAE model to generate ensemble candidates. In this figure we do not distinguish between inputs and outputs, showing that our ensembles can also work with a regular MAE, not just PHG-MAE. We do multiple random maskings which in turn result in a set of multiple reconstructions. Lastly, we combine them using an ensemble aggregation function, like averaging. In our experiments, we combine this technique with the PHG-MAE masking from Figure \ref{fig:mae-vs-io-mae} (right side) where inputs and outputs are clearly defined.}
    \label{fig:mae-test-time-ensembles}
\end{figure*}

We previously established that a neural hypergraph can be encoded with a single network using explicit masking. But what if we don't specify by hand these maskings and let the model learn the interdependencies between the modalities? The idea builds on a standard Masked Autoencoders (MAE) for Multi-modal multi-task learning (MTL), similar to \cite{bachmann2022multimae}. In Figure \ref{fig:mae-test-time-ensembles}, we present the core concept: the Multiple Randomly Masked Autoencoder Ensembles algorithm.

After the MAE model is trained, we can perform multiple inference steps through the model for the same input at test time. Each query results in a different random masking, which in turn results in a different random reconstruction. These reconstructions are then accumulated and combined through an aggregation function, as in ensemble learning. This approach can be applied to a regular Masked Autoencoder, however, in our case we combine this ensembling technique with the masking algorithm that defines input and output modalities. This is equivalent to the ensembles generated by multiple pathways of hypergraphs \cite{marcu2023self}. In our case, we simply average them out, but learned ensembles could also be employed as a future work. Assuming that the model lacks inherent randomness (e.g., softmax with temperature as in \cite{beeching2024scalingtesttimecompute}), then the number of valid candidates is limited to the full set of combinations of the masking strategy. For instance, with 4 input features, there is a total of $2^4-1=15$ valid candidates for ensembling. If the model incorporates randomness, the number of potential candidates grows immensely. Increasing the number of candidates stabilizes results by reducing prediction variance, a trend we also observe on our experiments on unseen scenes for UAVs. While reduced variance does not guarantee performance improvement, we observe both benefits in our experiments, consistent with other works on model ensembling \cite{mienye2022survey}.

\subsection{Intermediate-level Modalities for Ensemble Diversity, Richness and Generalization}
\label{subsec:intermediate-modalities}

In our experiments, which use the Dronescapes dataset, RGB is always an input modality as this is the default input given a new video of a new test scene. Conversely, semantic segmentation, depth estimation and camera normals estimation are always output modalities, thus always masked during training. However, this introduces a problem: if we were to only use the input and output modalities described above, we'd be unable to form ensembles and be restricted to a regular input-output prediction network. This is due to the fact that we only use RGB as input and predict three modalities with a single network using the masking strategy described earlier. In this context all the inputs (RGB) are always seen and all the three outputs are always masked. To solve this, we extract a set of intermediate modalities with a data-pipeline using pretrained expert models aligned with our output modalities.

\begin{figure*}[h]
    \centering
    \includegraphics[width=1\linewidth]{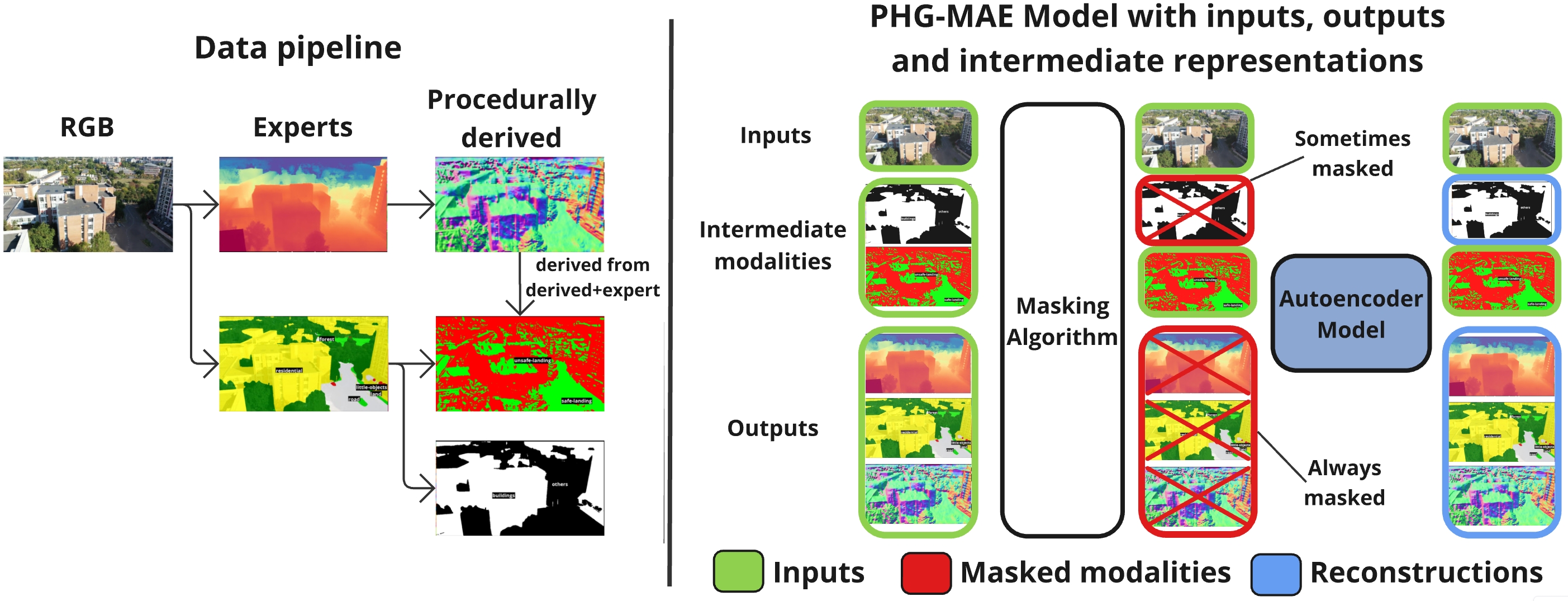}
    \caption{The Data-Pipeline and the PHG-MAE model. Left: The process of deriving modalities as pseudo-labels from pretrained experts using RGB only, followed by deriving new modalities from combinations of experts. Right: The integration of these experts \& generated modalities in the PHG-MAE semi-supervised training and inference procedure. Modalities are grouped in input, intermediate or output. Intermediate modalities are probabilistically masked and they drive the generation of ensembles at inference time. Different from other works, we mask at whole modality level, not at patch level. \\
    Note that in our experiments we generate up to 9 intermediate modalities out of 4 experts: 1 depth and 3 for semantic segmentation. However, this is in no way comprehensive and more modalities could be added in the future to further enhance the data. In the appendix we provide a full sample containing all of them.}
    \label{fig:data_pipeline}
\end{figure*}

We present the basic idea in Figure \ref{fig:data_pipeline}. First, we use the data-pipeline to extract expert data through pseudo-labels (i.e. semantic segmentation, unsupervised depth, optical flow, edges etc.). Then, based on these pretrained experts, we extract the intermediate modalities as simple derivations and combinations. Note that this is a simplified example and the exact list of experts and intermediate modalities that we derive are described in full detail in the Experiments section later on.

In the introduction we briefly discussed about task alignment from a data distribution point of view, and how some modalities are more similar than others, requiring different amounts of data or model parameters to learn the transformations from one to the other. Additionally, the task of semantic segmentation provides an extra challenge due to the arbitrarily defined set of fixed classes at different levels of abstraction and complexity. Higher-level classes like 'residential' areas are put together with more concrete ones, like 'water', 'land' or 'bike' and this leads to the models underperforming, especially in a multi-task learning setup like ours. We tackle this issue by creating intermediate modalities from pretrained experts, whose purpose is to act as a 'bridge' where high-level tasks (such as complex semantic segmentation) can be more easily learned from low-level raw modalities (such as RGB).

Moreover, these experts are trained on different datasets, which allows us to leverage their diverse core knowledge more effectively. Having diverse models is a requirement to achieve good results. Earlier, we introduced Multiple Randomly Masked Autoencoder Ensembles, which allows us to query the model at inference time multiple times after training in order to generate different predictions for the same input through random masking. This corresponds to making predictions with different hyperedges where, in our case, RGB is always present (input) and we get a random subset of intermediate modalities alongside it. Different subsets of modalities may enable different sub-networks that are more aligned with the output high level tasks. As these modalities are generated from pretrained data without any human supervision in an automated fashion, our PHG-MAE models are trained through a form of semi-supervised learning. For ensemble aggregation, in this work we do not use any learned ensembles, but we show that even the simple average improves the overall performance and consistency of the model, which we'll show in more detail in the Experiments section.

For our choice of intermediate modalities, we chose to use binary segmentations derived from different pretrained experts, namely Mask2Former variants trained on different datasets (COCO \cite{lin2014microsoft} and Mapillary \cite{neuhold2017mapillary}) or with different network architectures (Swin Transformer \cite{liu2021swin} and ResNet-50 \cite{he2016deep}). We hypothesize that these low to mid-level modalities are more robust to domain shifts and they can be seen as a missing link of Marr's tri-level hypothesis of information processing \cite{marr2010vision}. For instance, predicting binary maps for general concepts like "water" or "transportation vehicles" in UAV scenes is easier and less ambiguous than predicting fine-grained semantics, such as specific car models. Note that all these intermediate modalities are just recombinations of pretrained experts or other intermediate modalities, but they are in no way exhaustive. Further research in generating them in a more principled way would most certainly boost performance further. One useful principle that we followed is that the source must be a RGB frame and pretrained neural networks, so we can use any arbitrary videos to expand our dataset without any need for manual labels or expensive 3D reconstructions.

In summary, our Probabilistic hypergraphs using Masked Autoencoders model is defined by first creating two disjoint sets (inputs and outputs) combined into a single MAE-like model with full image-level random masking, boosted by intermediate modalities extracted from experts for improving test-time consensus via ensemble learning.

\subsection{Semi-supervised Learning and Model Distillation}

Model distillation \cite{hinton2015distilling} is a powerful technique used to compress the knowledge of a larger model, called a teacher, to a smaller model, called a student. This method allows smaller models to learn better representations using less data by mimicking the larger model instead of learning directly from data. In our case, we have an expensive data-pipeline followed by multiple random maskings to produce ensembles. This process takes a lot of time and slows down inference, however, through pseudo-labels and model distillation, we train very small models (150k, 430k parameters) with great performance. Moreover, we apply this method on new unseen data as well, which is a form of semi-supervised learning. We present our distillation results in Section \ref{subsec:experiments-distillation}.

\subsection{Temporal Consistency Metric}
\label{subsec:temporal-consistency}
We implement a temporal consistency metric for semantic segmentation, based on optical flow, following the work of \cite{marcu2023self}. The purpose of this metric is complementary to the frame-level performance metrics, such as Mean IoU which was used throughout this work. While our networks are trained on single frame predictions, deploying such models to the real world usually implies using a controller module that consumes these predictions, potentially adding latency. Thus, having temporal consistency metrics across time between multiple consecutive frames is desirable.

\begin{figure}[h]
    \centering
    \includegraphics[width=0.8\linewidth]{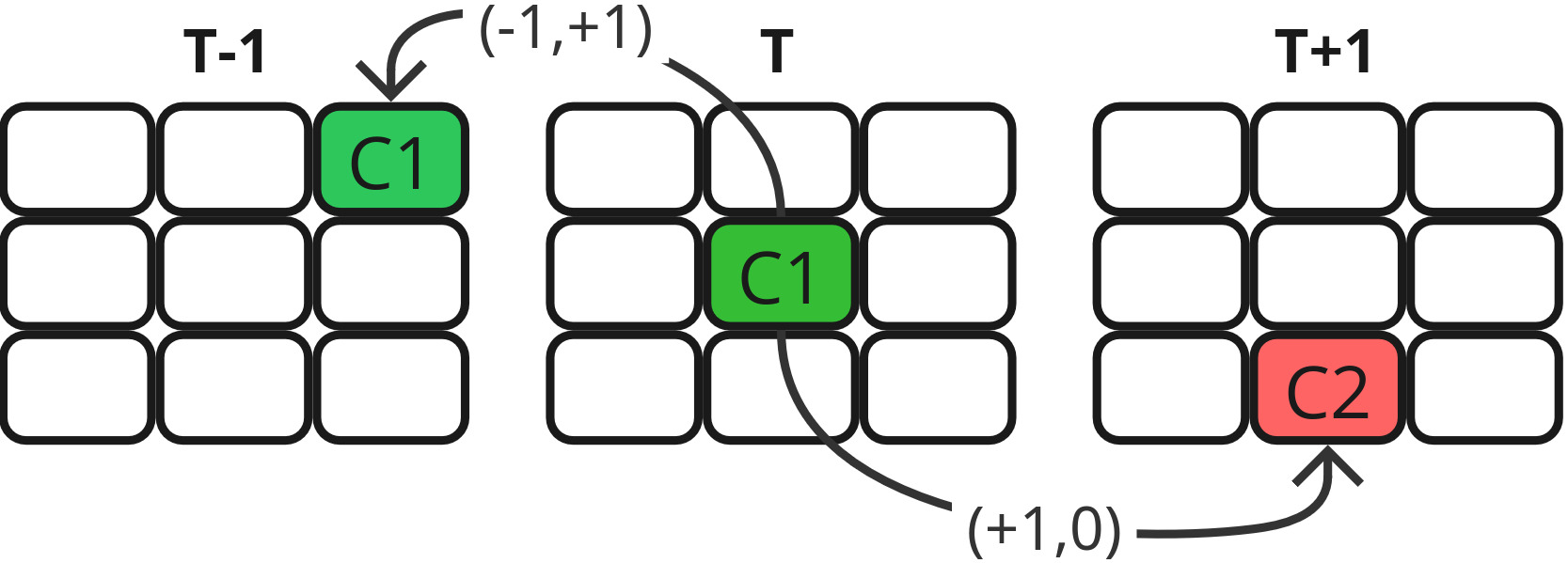}
    \caption{Temporal Consistency through Optical Flow. Example for a single pixel of a 3x3 frame resulting in a score of 0.5: the pixel in the previous image is consistent (same class), while the pixel on the right changed. This process is repeated for all the pixels in an image and averaged for a final per-frame score.}
    \label{fig:temporal-consistency-through-optical-flow}
\end{figure}

In order to do this, we measure the class consistency across the optical flow chain between consecutive frames, as seen in Figure \ref{fig:temporal-consistency-through-optical-flow}. For each frame $t$, we compute the forward and backward optical flows $t \rightarrow t-1$ and $t \rightarrow t+1$, using RAFT \cite{teed2020raft}. We then take the semantic segmentation prediction of this current frame and compare it to the prediction of the previous and next frames after warping it through optical flow. We have only three possibilities: no match with either frames, a single match or full match, to which we assign scores of 0, $1/2$ and 1. We then aggregate this score for each valid pixel of the current frame, dropping pixels where the warping would take us outside of the image or where occlusions might appear, indicated by optical flow. Finally, we take the average of all the frames in a video (except the first and the last one), resulting in a single consistency score across the entire video, which we multiply by 100, such that the metric can be interpreted as percentages.

\section{Experiments and Results}
\label{sec:experiments}

All our experiments were conducted on the \textit{Dronescapes} dataset benchmark for Aerial Scene Understanding, which we briefly summarize in the next section. All our experiments follow the same metrics as the original Dronescapes paper \cite{marcu2023self}: Weighted Mean Intersection over Union (Mean IoU in tables) for semantic segmentation and L2 * 100 for depth estimation and camera normals estimation. The exact formulas and class weights as presented in the original work are also described by us in the Appendix.

\textbf{Names and conventions:} throughout this section, our methods are referred to as \textit{PHG-MAE} followed by an optional suffix when appropriate: \textit{PHG-MAE-1Rand} refers to our method without ensembles (i.e. a pretrained model with a single randomly sampled edge), \textit{PHG-MAE-NRand} refers to our model followed by multiple random masking ensembles (i.e. n random edges plus ensemble) while \textit{PHG-MAE-Distil} refers to our distilled semantic segmentation model. Moreover, \textit{PHG-MAE-1All} refers to the MTL model equivalent to SafeUAV-MTL \cite{marcu2023self}, trained on our dataset variants. The suffix '1All' can be thought of a particular case where the masking algorithm uses all input modalities and masks all output modalities (i.e. one full hyperedge). In this light, the MAE-based 1Rand/NRand methods are particular edge samples of this 1All case.  Throughout this section, we primarily train a 4.4M-parameter model based on the original SafeUAV architecture, modifying it by increasing the number of input channels from 3 (just RGB) to include all channels from the additional modalities (inputs, outputs and intermediate). For ablation studies, we also train 1.1M, 430k and 150k-parameters models where appropriate.

Our experiments are organized as follows. First, in Section \ref{subsec:experiments-dataset-description}, we start by presenting the data and models setup that were used for training and evaluating our method, as well as aligning with prior work for comparison. The main focus of this work has been on semantic segmentation and, as such, we start in Section \ref{subsec:experiments-semantic-segmentation} with our results on the Dronescapes-Test dataset, followed by an analysis of the boost in performance provided by our Multiple Randomly Masked Autoencoder Ensembles method in Section \ref{subsec:experiments-ensembles}. In Section \ref{subsec:experiments-distillation}, we present our results using a parameter-efficient distillation process using as teacher the ensembled solution and as student a simple low parameter single-task CNN. Afterwards, in Section \ref{subsec:experiments-dataset-selection}, we provide a method for dataset pseudo-labels selection which improves the distillation results even further. We continue with Section \ref{subsec:experiments-temporal-consistency}, where we show that both our PHG-MAE as well as our distillation models outperform the large Mask2Former transformer model on a temporal consistency metric. In Section \ref{subsec:experiments-candidates-selection-algorithms}, we present our results on ensemble selection algorithms, showcasing the capabilities enabled by our PHG-MAE multiple random masking ensembles. In Section \ref{subsec:experiments-ablation-study} we present an ablation study regarding the number of intermediate modalities and the effect on ensembles. Finally, in Section \ref{subsec:experiments-mtl-results}, we present our results on Multi-Modal Multi-Task Learning (MTL), showcasing that our model is capable of predicting multiple tasks at once thus generalizing from semantic segmentation only to depth and camera normals estimation as well.

\subsection{Data and Learning}
\label{subsec:experiments-dataset-description}

\subsubsection{Dronescapes Dataset}
\begin{figure*}[h]
    \centering
    \includegraphics[width=1\linewidth]{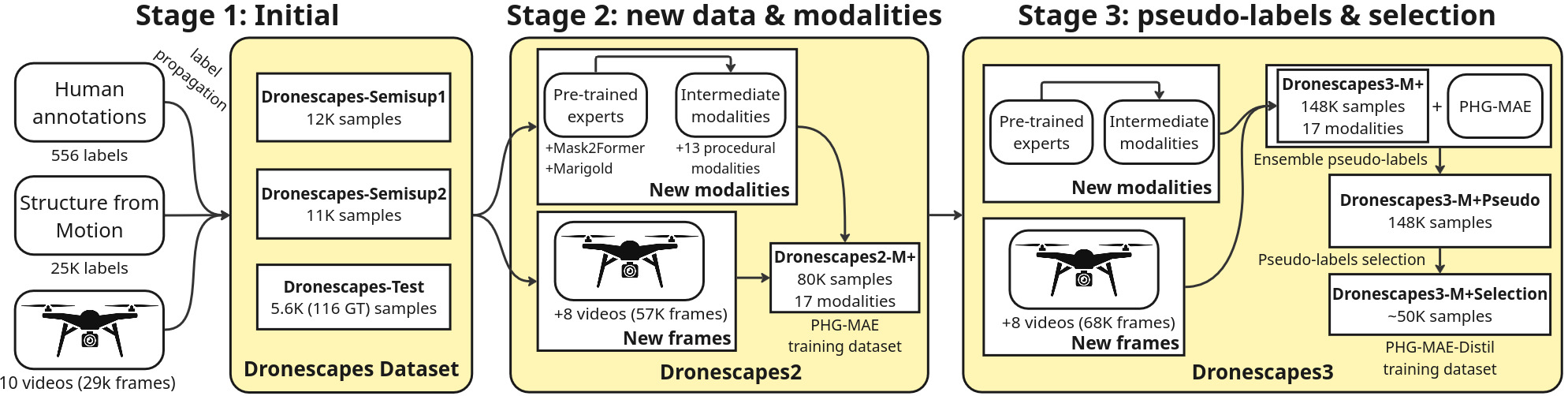}
    \caption{Dronescapes dataset genesis. Stage 1 refers to the original dataset generation \cite{marcu2023self}, which uses Structure from Motion \cite{alicevision2021} to generate metric depth and camera normals, as well as human annotation plus label propagation \cite{marcu2020semantics} for semantic segmentation. Stages 2 and 3 are introduced in this work, each new stage adding 8 new UAV videos. Stage 2 is a fully-automated process designed to train the PHG-MAE model starting from RGB frames only. It is split in two independent axes: new frames (57K new frames from 8 new videos) and new modalities. We use four pretrained experts: 3 Mask2Former variants \cite{cheng2022masked} for semantic segmentation and Marigold \cite{ke2024repurposing} for depth and camera normals. Based on them, we add 13 new modalities (4 from experts and 9 procedurally-generated), following the process from Figure \ref{fig:data_pipeline}. This stage results in the Dronescapes2-M+ dataset, which is used to train our PHG-MAE model. In Stage 3 we first repeat the steps of adding 8 new videos and generating modalities, followed by producing ensemble-based pseudo-labels for semantic segmentation based on the consensus of our model. Finally, we apply a consistency-based selection algorithm to filter out noisy pseudo-labels, resulting in the Dronescapes3-M+Selection dataset which is used to train our efficient distillation model: PHG-MAE-Distil.}
    \label{fig:dronescapes-genesis}
\end{figure*}

We start with brief information about the original dataset, its variants, number of scenes and data points. Then, we define our contributions to it done in order to improve the generalization: extending it with more scenes and augmenting it with pretrained experts and intermediate modalities. For a thorough technical documentation and reproduction steps of both the original dataset as well as our extended variants, you can visit the official website: \url{https://sites.google.com/view/dronescapes-dataset}. All the variants of the Dronescapes dataset, both original and our extensions, with links between them are presented in Figure \ref{fig:dronescapes-genesis}. Moreover, Table \ref{tb:dronescapes-variants} provides a summary of the size and number of modalities to contextualize the experiments shown in later sections.

\begin{table*}[h]
\centering
\scalebox{0.84}{
\begin{tabular}{|l c c c l|}
    \hline
    Name & Data Points & I/O         & UAV scenes & Description \\
         & (GT labels) & Modalities  &            & \\
    \hline
    Dronescapes-Semisup1 \cite{marcu2023self} & 12K (233) & 5/3 & 7 & First set of pseudo-labels from analytic SfM \\
    & & & & and label propagation via SegProp \cite{marcu2020semantics}. \\
    \hline
    Dronescapes-Semisup2 \cite{marcu2023self} & 11K (207) & 5/3 & 7 & Second set of pseudo-labels. \\
    \hline
    Dronescapes-Test \cite{marcu2023self} & 5.6K (116) & 5/3 & 3 & Original test set. All benchmarks are run on it. \\
    \hline
    \multicolumn{5}{|c|}{\textbf{Our work below}} \\
    \hline
    Dronescapes-Semisup1-M+ & 12K (233) & 14/3 & 7 & Stage 2: New experts and modalities \\
    \hline
    Dronescapes2 & 80K (440) & 1/3 & 15 & Stage 2: 8 new UAV scenes (+57K samples) \\
    & & & & plus Dronescapes-Semisup1+2 \\
    \hline
    Dronescapes2-M+ & 80K (440) & 14/3 & 15 & Result of Stage 2: 8 new scenes, experts and \\
     & & & & modalities. Used to train PHG-MAE. \\
    \hline
    Dronescapes3-M+ & 148K (440) & 14/3 & 23 & Result of Stage 3: Another 8 new scenes, \\
     & & & & experts and modalities. Used for distillation. \\
    \hline
\end{tabular}
} 
\caption{Dronescapes dataset variations and stats. Numbers in parentheses represent the semantic human-annotated data. 'I/O Modalities' refer to inputs and outputs. The first three rows correspond to the three initial sets introduced in \cite{marcu2023self}: first semi-supervised set ($\sim$12K samples \& 233 human-annotated), the second semi-supervised set ($\sim$11K samples \& 207 human-annotated) and the test set ($\sim$ 5.6K samples \& 116 human-annotated). We extend the dataset in two dimensions: number of modalities (row 4) and frames (row 5). Our final dataset combines the two approaches resulting in the Dronescapes2-M+ (row 6) and Dronescapes3-M+ (row 7) variants which contain up to 148K samples and 17 modalities.}
\label{tb:dronescapes-variants}
\end{table*}

The dataset defines three tasks: semantic segmentation, depth estimation and camera normals estimation. The evaluation is typically conducted on the 116 test frames where both human annotations and SfM reconstructions are available. Semantic segmentation maps are human-annotated, with label propagation \cite{marcu2020semantics} used to interpolate missing frames. Depth and camera normals were generated from raw videos and GPS data using a Structure-from-Motion (SfM) tool \cite{alicevision2021}. It is worth noting that Dronescapes-Test contains 5.6K samples, but SfM-based depth and camera normals ground truth are available for only one scene due to unreliable reconstructions in the other two. Consequently, these two modalities can only be evaluated on 1.4K samples, of which only 36 include human-annotated semantic maps.

\subsubsection{Extended Dataset: New Scenes and Modalities}
\label{subsubsec:extending-the-dataset}
We extend the Dronescapes dataset by two complementary dimensions: 1) adding new UAV videos to increase diversity and 2) enriching the representations based on pretrained experts by introducing new procedurally-derived modalities. Our convention is as follows: when we add new scenes, we increment the Dronescapes number (i.e. Dronescapes2, 3 etc.). When we add new modalities or perform additional operations (i.e. filtering), we add a proper suffix (i.e. Dronescapes2-M+). We validate all our additions on the Dronescapes-Test dataset as the main guiding tool. Note that these frames are never seen by any of our trained models avoiding data leakage, as we hope that our methods are useful for generalization on aerial image understanding tasks and beyond, not just benchmarks improvements.

We extend the dataset in two stages on top of the original work of \cite{marcu2023self}: Stage 2 focuses on new data and new modalities. Stage 3 also contains new data and modalities, but goes beyond that in the realm of generating and selecting pseudo-labels for efficient model distillation and iterative semi-supervised learning \cite{leordeanu2021semi}.

\textbf{Stage 2} We start by adding 8 new videos sourced from the internet, adding a total of 57K new frames, which we add on top of the Dronescapes-Semisup1+2 combined dataset, totaling at around 80K samples. We call this dataset simply \textit{Dronescapes2} which contains a total of 15 UAV scenes. These additions do not include human annotations; instead, the three output tasks are automatically generated using the pretrained experts. The new videos are chosen to increase the diversity of the data as much as possible, by incorporating data from different regions, countries, seasons, weather conditions and even cameras, while maintaining a fully automated setup to allow further extensions. For semantic segmentation, we use Mask2Former \cite{cheng2022masked}, specifically the median of three released pretrained checkpoints: \textit{49189528\_1}, \textit{49189528\_0}, and \textit{47429163\_0}, after applying a task-based mapping from the source classes (e.g. from Mapillary \cite{neuhold2017mapillary} and COCO \cite{lin2014microsoft}) to the 8 Dronescapes classes. For depth estimation, we use Marigold \cite{ke2024repurposing} (unscaled depth) and derive camera normals using an SVD-based algorithm on the expert depth map, following \cite{hartley2003multiple}. These three modalities are closely aligned with the output tasks of the Dronescapes-Test benchmark. Moreover, we adapt their raw distribution (mean and standard deviation), using the statistics derived from the Dronescapes-Semisup1 such that they are suitable to be used as ground truth during training.

On the other hand, based on the pretrained experts, we introduce up to 13 new procedurally-generated modalities, following the process described in Figure \ref{fig:data_pipeline}. For comparison purposes, we initially validate this method on the original dataset, creating \textit{Dronescapes-Semisup1-M+}, where the M+ suffix stands for "newly added modalities". The set of intermediate modalities that we derive are: \textit{camera normals from depth}, \textit{vegetation}, \textit{sky-and-water}, \textit{containing}, \textit{transportation}, \textit{buildings (all types and nearby only)} and \textit{safe-landing (geometric only and geometric + semantic)}. In total, we have four new experts (3 Mask2Former variants) and nine new intermediate modalities. The exact mappings and formulas are described in the Appendix. Note that these modalities are hand-crafted and they are neither final, nor perfect. We aim to automate this process later on to discover or filter intermediate modalities based on performance metrics.

\textbf{Stage 3} This stage further extends the data by adding yet another 8 videos as well as deriving new intermediate modalities, totaling at 148K frames across 17 modalities, raw, experts and procedurally derived. We call this \textbf{Dronescapes3-M+}. This stage also focuses on generating pseudo-labels for efficient teacher-student distillation. We take our best model from Stage 2, generate pseudo-labels for semantic segmentation followed by training a simple single-task RGB $\rightarrow$ semantic distilled network on these frames. This stage also focuses on efficient pseudo-labels selection.

\subsubsection{Models and Training Description}
\label{subsec:experiments-models-description}

\begin{figure*}[h]
    \centering
    \includegraphics[width=1\linewidth]{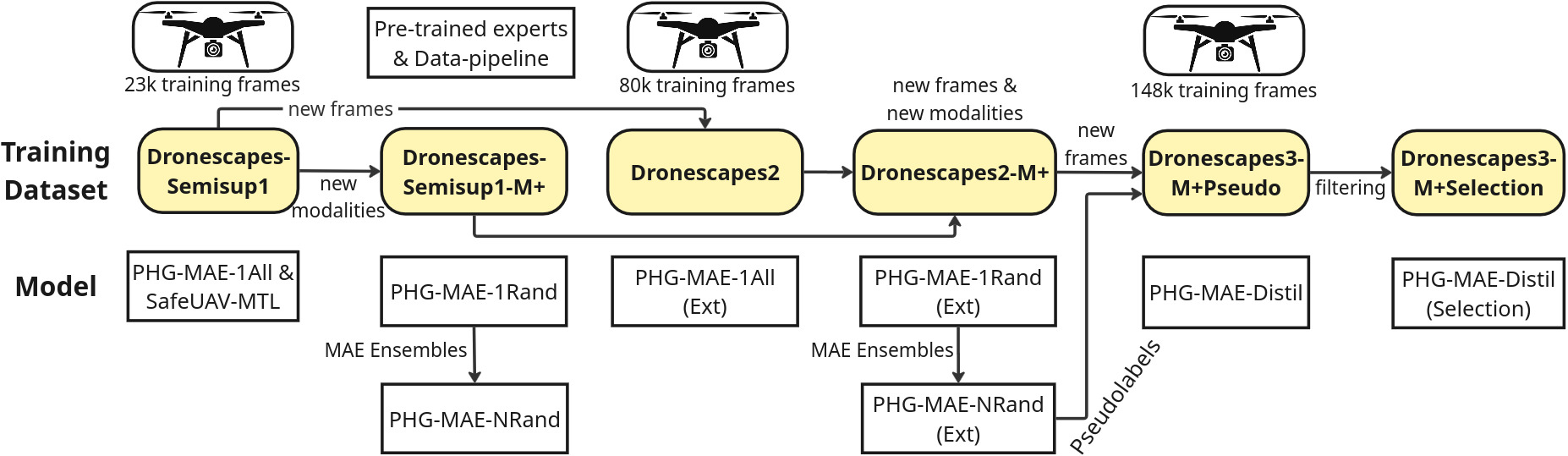}
    \caption{Models genesis. We start with the original \textit{Dronescapes-Semisup1} dataset and the original SafeUAV-MTL models \cite{marcu2023self}. On this dataset we train our PHG-MAE-1All equivalent model for a fair comparison. Then, using the pretrained experts and the procedurally generated intermediate modalities, we generate the Dronescapes-Semisup1-M+ dataset, which is used to train the first PHG-MAE model. Complementary, we extend the initial dataset with 8 new UAV videos resulting in the \textit{Dronescapes2} variant (RGB and 3 output modalities only) on which we train our PHG-MAE-1All (Ext) model. We combine the two by extracting experts and intermediate modalities on the new scenes resulting in \textit{Dronescapes2-M+}, the dataset used to train our PHG-MAE (Ext) model. Using this model, we generate pseudo-labels for semantic segmentation resulting in \textit{Dronescapes3-M+Pseudo}. Furthermore, we filter these pseudo-labels using a consistency-based algorithm with feedback from our PHG-MAE model. These datasets are used to train our lightweight PHG-MAE-Distil models.}
    \label{fig:models-genesis}
\end{figure*}

This subsection aims at making the reader more familiar with the names of the models and the experiments done in the rest of the section, as well as the datasets used for each. In Figure \ref{fig:models-genesis}, we present a visual summary of the models and how they relate to each other.

As all the data is concatenated at channel level before entering the CNN model, the modalities share direct information in the first layer, learning interdependent relationships. The masked modalities are zeroed out, but they are also directly associated in the output layer, so the model learns to associate the missing information through backpropagation based on the reconstruction loss. This masking model was described in Section \ref{subsec:phg-modeling-hg-with-mae}. We train all our models on a single server with 8 NVIDIA RTX 2080 Ti GPUs using DDP \cite{li2020pytorch}, which are consumer grade GPUs and three generations old at the time of this writing. Our research was also focused on efficient model training methods, without requiring access to hundreds or thousands of server grade GPUs. We trained all the models using images at a resolution of 540x960 and a random cropping augmentation + rescaling method (up to 50\% of the image), using the PyTorch framework and the AdamW optimizer \cite{loshchilov2017fixing} following a Cyclic LR scheduler. We train for up to 150 epochs and pick the best checkpoint based on the validation set. We release our training/evaluation code and checkpoints for reproducibility, alongside an inference notebook for online inference on any input videos. In the Appendix we also provide a table with the training durations on various model sizes and dataset variants.

\subsection{Results: Multiple Randomly Masked Autoencoder Ensembles}
\label{subsec:experiments-ensembles}

\begin{figure}[h]
\centering
\includegraphics[width=1\linewidth]{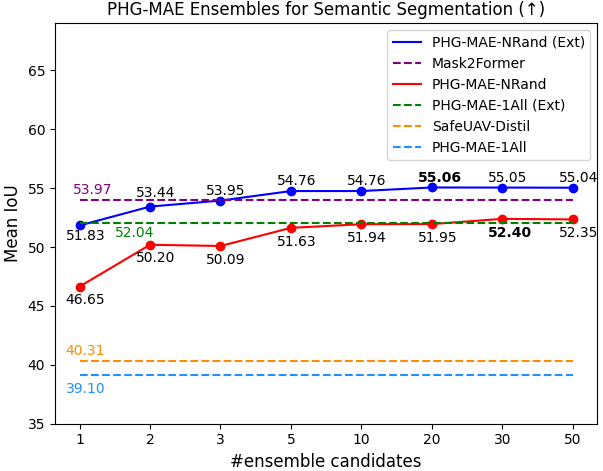}
\caption{Semantic segmentation results with ensembles against the baselines. The (Ext) variants refer to the models trained on the extended dataset Dronescapes2-M+ (80K samples), while the one without are trained on the original frames of Dronescapes-Semisup1-M+ (12K samples). Mask2Former \cite{cheng2022masked} (violet line) is one of the current state-of-the-art transformer models for semantic segmentation. The PHG-MAE-1All model (green \& cyan lines) is a CNN-based model akin to SafeUAV-Distil \cite {marcu2023self} (orange line) without any inference-time ensembles. For the ensemble models (-NRand), we provide results varying the number of candidates (N) from 1 up to 50 showcasing the effectiveness of the method as the number of candidates increases.}
\label{fig:results-sseg-ensembles}
\end{figure}

In this section, we present results using our MAE-based ensemble algorithm introduced in Section \ref{subsec:random-masking-ensembles} on the task of semantic segmentation on the Dronescapes-Test benchmark. These results are presented in Figure \ref{fig:results-sseg-ensembles}.

As baselines, we provide a couple of "static" models, whose performance cannot be improved at inference-time by performing ensembles. Our baseline PHG-MAE-1All trained only on the Dronescapes-Semisup1 (12K samples) performs at 39.10, while the SafeUAV-Distil variant (23K samples) performs at a score of 40.31. Moving on, our second baseline (green line), which is simply trained on the extended Dronescapes2 dataset (80K samples), without any extra modalities besides RGB and the three output tasks performs at a score of 52.04. For a state-of-the-art model, we use Mask2Former \cite{cheng2022masked}, a 216M-parameters model trained on the Mapillary dataset \cite{neuhold2017mapillary}, which sits at 53.97.

We now move to our PHG-MAE-NRand models, which use the inference-time ensemble algorithm. We vary our N from 1 candidate all the way to 50 candidates to contextualize the boost offered by the method. For using a single sample, which we call PHG-MAE-1Rand, the models achieve scores of 46.65 (12K samples) and 51.83 (80K samples). Interestingly, when using only 12K samples this is well above the 39.10 score of -1All, however, for 80K samples this is below the static score of 52.04. We conclude that the added intermediate modalities offer a relevant signal on low-data situations, leading to significant boosts in performance. As we increase the number of candidates, we observe large performance gains obtained immediately, with the numbers continuing to grow all the way towards 20 candidates. The -NRand variant trained on only 12K samples outperforms the -1All variant trained on 80K once it reaches 30 candidates. This means that while increasing the number of frames in our dataset can provide good initial boosts, increasing the number of modalities has a greater benefit with the addition of inference-time ensembles. This is one of the reasons why we split the Stage 2 in Section \ref{subsubsec:extending-the-dataset} into two independent axes: extending the number of frames and extending the number of modalities, as both provide similar and independent boosts. When put together the fully-extended dataset provides state-of-the-art results, going from 51.83 (-1Rand) all the way to a score of 55.06 at N=20 candidates.

\begin{figure*}[h]
    \centering
    \includegraphics[width=1\linewidth]{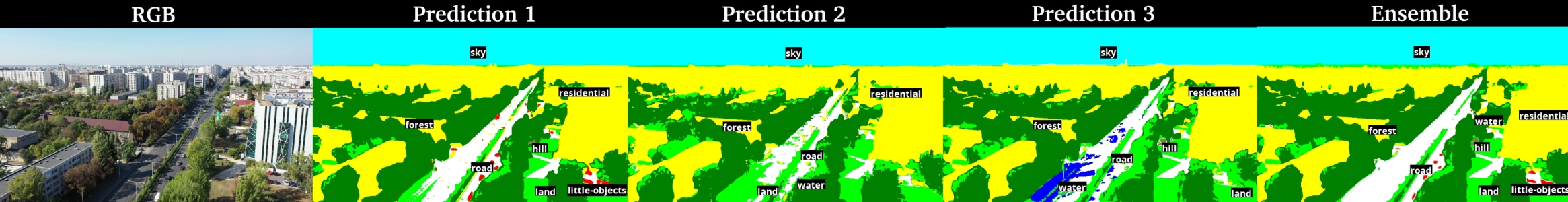}
    \caption{Three independent predictions on an unseen scene, each of them with various flaws (i.e. little objects not seen, water on roads etc.) while the aggregated ensemble (over 30 candidates here) provides a consistent result that discards all the little errors and keeps the most likely predictions agreed upon by each candidate on average.}
    \label{fig:results-test-time-ensembles-qualitative2}
\end{figure*}

\begin{figure*}[h]
    \centering
    \includegraphics[width=1\linewidth]{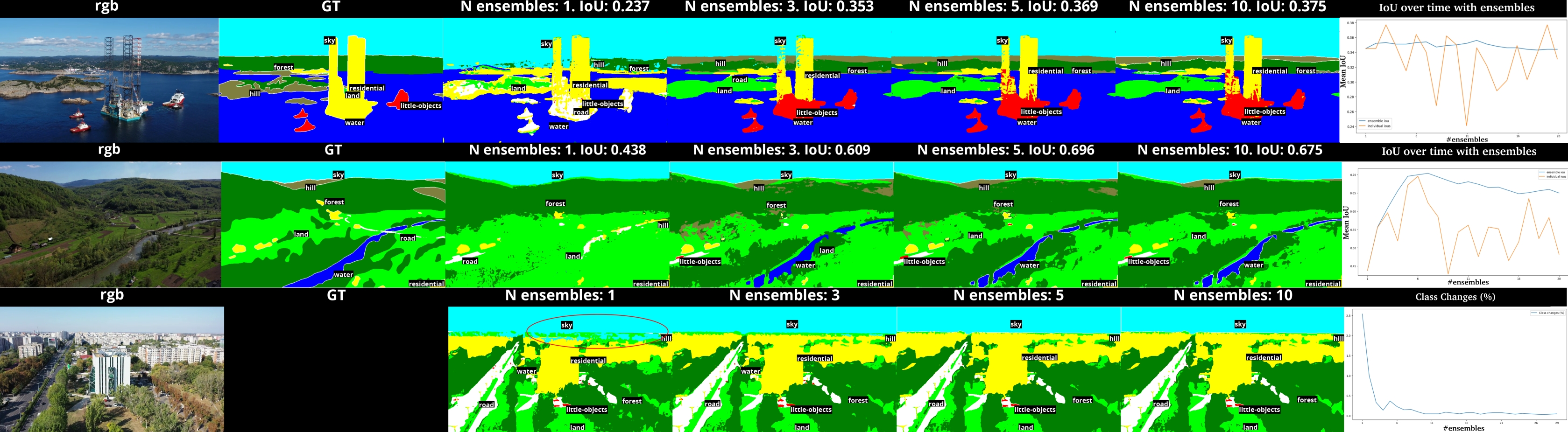}
    \caption{Multiple Randomly Masked Autoencoder Ensembles qualitative results on unseen testing scenes. First two rows are from Dronescapes-Test, enabling an analysis of ensembles performance w.r.t IoU score. Last row is from a new video showcasing the stability when adding more candidates. We highlight a distant area where a single prediction has noise (uncertainty) that is reduced by ensembling. Last column showcases the convergence of ensemble learning towards a stable solution.}
    \label{fig:results-test-time-ensembles-qualitative}
\end{figure*}

\textbf{Qualitative Results.} In Figure \ref{fig:results-test-time-ensembles-qualitative2} we show three independent predictions followed by an ensemble result where we can see that while each sampled candidate has its own flaws, when put together they manage to cancel each other. Moreover, in Figure \ref{fig:results-test-time-ensembles-qualitative} we see three more qualitative results. The first two rows are samples from the annotated test set. They include the ground truth human-annotated semantic map, allowing step-by-step performance analysis of ensemble candidates. We notice a high IoU variance for each candidate, likely due to the ensemble's internal structure, where some modalities align better with the GT semantic map. However, the aggregated ensemble solution tends to be more stable and often better. The last row is from an unseen video, so it doesn't have any ground truth semantic map. However, we can measure prediction changes as more candidates are added (last column). We observe the model converges to a solution, with fewer than 1\% pixel class changes after about 10 candidates.

\subsection{Teacher-Student Distillation from PHG-MAE Models}
\label{subsec:experiments-distillation}

Due to the multi-modal multi-task nature of our PHG-MAE model, it can become very complex and slow to run on new data. To obtain a prediction, one must run the data-pipeline on each RGB frame, doing inference on all the pretrained experts, followed by re-generating all the intermediate modalities. On top of that, if we want to do PHG-MAE ensembles (-NRand variant), we must run the model N times and average all these independent predictions together. The more input modalities the model has, the more time is required to compute the input data. Furthermore, the memory footprint is determined by the largest expert model, in our case Mask2Former. While this is not an issue in offline settings, deploying such a complex model becomes challenging in practice. For this purpose, we distill our best performing semantic segmentation model, resulting in a model that only needs RGB inputs. Effectively, this distillation process compresses the pretrained PHG-MAE model, the test-time ensembles, as well as the data-pipeline into the weights of the new model. The results can be seen in Table \ref{tb:results-distillation-semantic}.

\begin{table*}[h]
\centering
\scalebox{0.94} {
\begin{tabular}{|c c c c c|}
    \hline
    Model & Parameters & Training   & Mean IoU $\uparrow$ & Runtime (s) $\downarrow$ \\
          &            & Dataset     &                     &                          \\
    \hline
    \textbf{PHG-MAE-NRand} & \textbf{4.4M} & \textbf{Dronescapes2-M+} & \textbf{55.06 ± 0.09} & \textbf{78.9} \\
    \hline
    \underline{PHG-MAE-Distil} & \underline{4.4M} & \underline{Dronescapes3-M+Pseudo} & \underline{55.05} & \underline{0.064} \\
    \hline
    PHG-MAE-Distil & 430k & Dronescapes3-M+Pseudo & 54.94 & 0.054 \\
    \hline
    PHG-MAE-Distil & 1.1M & Dronescapes3-M+Pseudo & 54.30 & 0.058 \\
    \hline
    Mask2Former \cite{cheng2022masked} & 216M & Mapillary* \cite{neuhold2017mapillary} & 53.97 & 0.79 \\
    \hline
    PHG-MAE-Distil & 150k & Dronescapes3-M+Pseudo & 53.32 & 0.052 \\
    \hline
    SafeUAV-Distil \cite{marcu2023self} & 1.1M & Dronescapes-Semisup1+2-Pseudo & 40.31 & 0.052 \\
    \hline
\end{tabular}
} 
\caption{Model distillation for semantic segmentation using the ensemble pseudo-labels of the PHG-MAE-NRand teacher model. We observe little to no degradation in performance during model distillation, especially for the 4.4M model, while reaching an almost 1000x runtime speed-up, as these models require only RGB images. Furthermore, the models offer competitive performance against Mask2Former, a model that is 2 to 3 orders of magnitude larger. We also compare against SafeUAV-Distil \cite{marcu2023self} trained with pseudo-labels generated by their neural hypergraph model that also includes ensembles, though they are fixed and not probabilistic, like ours.}
\label{tb:results-distillation-semantic}
\end{table*}

To generate the pseudo-labels we set N=20 in our PHG-MAE-NRand model, as this provided the best results in Figure \ref{fig:results-sseg-ensembles}. We generate pseudo-labels for all the frames of Stage 3 as introduced in our Dronescapes dataset extension in Section \ref{subsubsec:extending-the-dataset} resulting in the \textit{Dronescapes3-M+Pseudo} which contains 148K RGB and semantic segmentation samples. We train the student models only using KL divergence on the logits of the teacher model, thus the student model never sees the original ground truth label or the expert's pseudo-label, only RGB as input and the teacher's output logits. We observe that the student model reaches within 0.01 Mean IoU points of the teacher for the similarly-sized 4.4M student model. We see similar performance for the 1.1M and 430k-parameters sized models, with scores of 54.30 and 54.94. Both these models outperform the Mask2Former baseline model, having a 200 $\sim$ 500x reduction in parameters and thus memory footprint. Our smallest model, which has only 150k parameters reaches a good performance as well, at 53.32, just 0.65 Mean IoU points from the large transformer.

We also report the inference runtime on a new test image from scratch on an RTX 2080 Ti, including all data preprocessing, after loading the model in memory. On average, it takes about 74.4 seconds to run the data-pipeline on a single frame plus yet another 4.5 seconds to gather the N=20 independent predictions and average them together. Note, that this includes loading Mask2Former three times and Marigold, both being big Transformer models, alongside generating all intermediate modalities. On the other hand, the distilled models are designed specifically for efficient inference, by having a single input (RGB) and a single output (semantic segmentation) which are suitable for real-time semantic segmentation use-cases.

\subsubsection{Automatic Pseudo-labels Selection for Teacher-Student Distillation}
\label{subsec:experiments-dataset-selection}

In the previous subsection, we've provided a comprehensive set of experiments for doing unsupervised model distillation, leading to great results which almost match the performance of the teacher PHG-MAE-NRand model. However, for all these experiments, we used the entire Dronescapes3-M+Pseudo (148K) pseudo-labels, regardless of their quality. In this subsection we want to filter out some of these pseudo-labels based on unsupervised metrics that act as a proxy for performance, which could be beneficial in three ways: first, it would lead to better performance as the model will not learn from uncertain or noisy labels; second, it could potentially boost the performance of the student even beyond that of the teacher on the Dronescapes-Test benchmark for semantic segmentation. And third, using less data will also reduce the training duration.

\begin{figure}[h]
    \centering
    \includegraphics[width=1\linewidth]{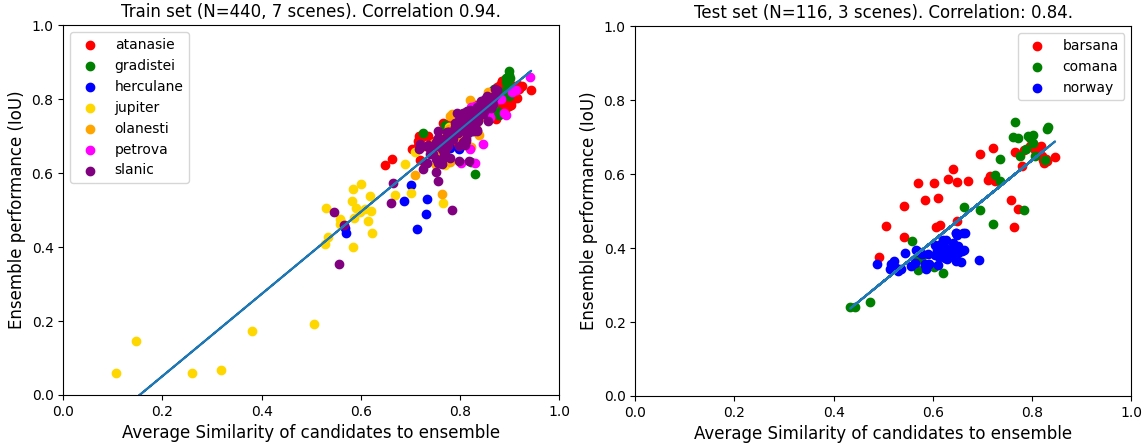}
    \caption{Scatter plot of the candidates' average similarity and the performance of the ensemble. We provide results on both subsets of the Dronescapes dataset where human annotation exists. The x axis refers to the average similarity of each of the 50 exported candidates to the average ensemble they produce. The y axis refers to the performance of the average ensemble to the ground truth map. We observe a positive correlation in both sets, meaning that the average similarity can be used as a proxy performance metric for pseudo-labels selection on unlabeled data.}
    \label{fig:experiments-dataset-selection-scatter}
\end{figure}

We start by making use of all the human-labeled frames on both the train and the test set, which we'll call simply GT (ground truth). Then, for each of these frames, we generate 50 unique candidates by doing 50 independent passes through the PHG-MAE-NRand model. Finally, we average these candidates to create a random ensemble. Thus, for each frame we have three items: the GT, 50 candidates and the averaged ensemble. We then compute two numbers: the average similarity of the candidates with regard to the ensemble (1) and the performance of the ensemble with regard to the GT (2). What we want to observe is whether there is any correlation between the internal consistency of the candidates and the overall group's performance. For both the similarity of each candidate against the ensemble we use the very same formula: the Weighted Mean IoU used throughout this work. Finally, we average the 50 similarities together resulting in a single number per frame. The scatter plot of this experiment on all the samples which contain human annotations can be seen in Figure \ref{fig:experiments-dataset-selection-scatter}.

\begin{figure}[h]
    \centering
    \includegraphics[width=1\linewidth]{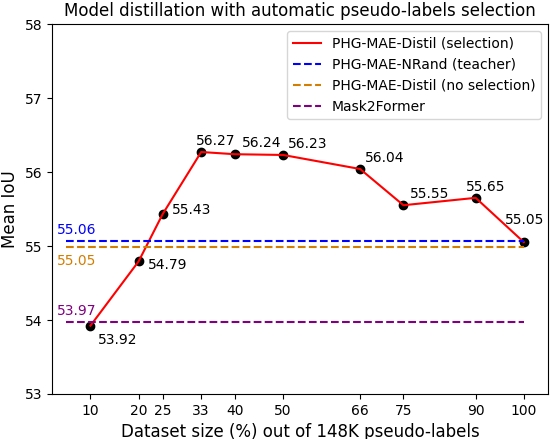}
    \caption{Semantic segmentation unsupervised distillation results using automatic pseudo-labels selection based on the average similarities of the candidates against the ensemble using the Dronescapes3-M+Pseudo dataset (148K samples). The baseline teacher (blue line) was trained on Dronescapes2-M+ dataset (80K samples), while the regular distillation (orange line) was trained on the entire set of 148K pseudo-labels. We also present Mask2former (violet line) as a point of reference. When applying pseudo-labels selection, we manage to improve the teacher's performance by 0.37 IoU points, from 55.06 to 55.43 using only 25\% of the selected pseudo-labels. Then, by using only a third of the pseudo-labels, we jump to a score of 56.27, more than 1 point above the teacher, showcasing the effectiveness of the method. This result more or less stagnates up until using 66\% of the dataset, after which it drops below a score of 56 again.}
    \label{fig:results-distillation-dataset-pseudo-labels-selection}
\end{figure}

We observe that there is a strong correlation between the two axes: the greater the average similarity, the greater the performance as well. This makes intuitive sense, as it means that many of the random candidates of the PHG-MAE-NRand model are in agreement with each other. A second observation is that each scene tends to have its own distribution as these data points cluster next to each other. For example, the Norway scene on the test set or the Jupiter scene on the original training set have an overall lower performance and similarity compared to the other ones. Using this positive feedback on the GT labels, we compute the similarities of all the 148K pseudo-labels generated from 50 random candidates on the Dronescapes3-M+Pseudo dataset. Then, we aim to pick only the top N\% candidates based on the thresholds provided by the similarity of the candidates to the ensemble as a proxy metric. We train only the 4.4M model, which we compare with the variants presented earlier in Table \ref{tb:results-distillation-semantic} where we used the entire dataset for pseudo-labels without any selection. The results can be seen in Figure \ref{fig:results-distillation-dataset-pseudo-labels-selection}.

We tried two ways of picking the top N\% of the pseudo-labels: global threshold and a per-scene threshold. It turns out that each scene has its own distribution of similarities based on the complexity of the scene (i.e. see the scatter plot where the points of the same colors tend to cluster next to each other). Using a global threshold resulted in some scenes being completely removed from the pool of pseudo-labels, which in turn resulted in worse performance. For N=25 (using 25\% of the pseudo-labels), the global threshold yielded a score of 54.01, while using a per-scene threshold yielded a score of 55.43. Based on this feedback, we only used the second threshold for the rest of the experiments which provides a good balance between pseudo-labels consistency and dataset diversity. We provide the full dataset similarities distribution across all the scenes in the appendix.

We observe that by using only a quarter of the dataset, we manage to improve the results of the teacher from 55.06 to 55.43 and by using a third of the dataset ($\sim$49K samples), we improve the results by more than one point, from 55.06 to 56.27. This result is remarkable and provides a good insight that quantity is not the only thing that matters, and we need better ways to filter out noisy or data that is not diverse. As we only use video data as source for the Dronescapes dataset, the number of frames is large but also highly redundant, thus automatic pseudo-labels selection methods such as this one can be of great benefit.

\subsection{Results: Temporal Consistency}
\label{subsec:experiments-temporal-consistency}

This subsection presents the temporal consistency metric scores using the method presented in Section \ref{subsec:temporal-consistency}. We compared three models on a diverse set of videos: Our PHG-MAE-NRand model, our PHG-MAE-Distil (4.4M) model and the Mask2Former \cite{cheng2022masked} baseline model. The results can be seen in Table \ref{tb:temporal-consistency-scores}.

\begin{table}[h]
    \centering
    \scalebox{0.96}{
    \begin{tabular}{|c c c c c|}
        \hline
                         & \multicolumn{4}{c|}{\textbf{Testing dataset}} \\
        \hline
        Model            & (1) & (2)  & (3) & (4) \\
        \hline
        \textbf{PHG-MAE-Distil}   & \textbf{98.41}   & \textbf{98.28}   & \textbf{98.21}    & \textbf{98.15}   \\
        \hline
        \underline{PHG-MAE-NRand}      & \underline{96.03}            & \underline{97.03}            & \underline{96.83}             & \underline{97.26}            \\
        \hline
        Mask2Former      & 95.71            & 93.23            & 95.19             & 96.79            \\
        \hline
    \end{tabular}
    } 
\caption{Consistency scores of our two best models against Mask2Former on Dronescapes-Test (1), Dronescapes2-M+ (2) \& Dronescapes3-M+ (3) (8 videos in each case), as well as two completely new videos from the internet (4). Higher is better.}
\label{tb:temporal-consistency-scores}
\end{table}

As this metric computation is completely unsupervised, we can use any semantic segmentation predictions as long as we have access to the optical flows of each frame. While our PHG-MAE model, ensembled across 30 candidates, provides better results than Mask2Former in terms of temporal consistency, the 4.4M distilled version shows a better overall score. This further implies that these distilled models can be used for real autonomous robotics applications showcasing good performance and great consistency.

\subsection{Analysis of Ensemble Selection Algorithms for Semantic Segmentation}
\label{subsec:experiments-candidates-selection-algorithms}

In this section we present a brief set of experiments aimed at understanding and improving the ensemble candidates selection mechanism from our PHG-MAE model based on the quality of these candidate hyperedges. Throughout the previous experiments, we used the algorithm presented in Figure \ref{fig:mae-test-time-ensembles} where each sampled hyperedge contains a set of intermediate modalities (nodes) each of which has a 50\% chance of being masked, following the same protocol used at training time. Then, through multiple random maskings, all the sampled hyperedges act as candidates in the ensemble and are later aggregated using the simple average. We showed earlier in Figure \ref{fig:results-sseg-ensembles}, that even with this simple method, increasing the number of candidates yields better results without any re-training. However, we ask the question: \textbf{can we do better than a fixed candidates selection algorithm followed by simple average?} Can we dynamically select the right candidates and discard noisy ones? Answering these questions would potentially unlock even greater performance gains at inference-time compared to a fixed protocol of masking and averaging.

\begin{figure}[h]
    \centering
    \includegraphics[width=1\linewidth]{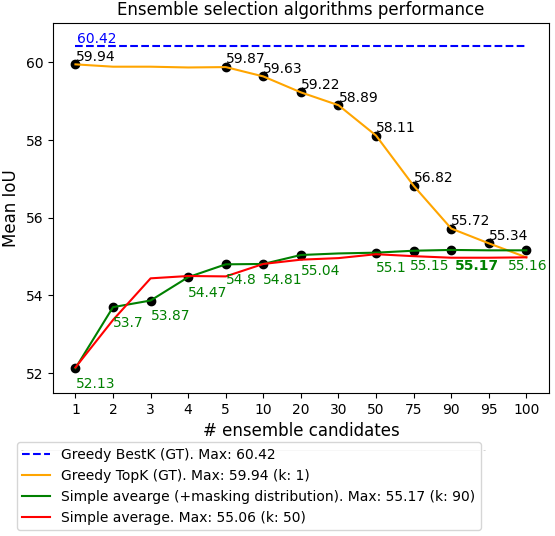}
    \caption{Ensemble selection using an oracle. We compare the baseline Simple average presented throughout the paper against the variant where we use the masking distribution, which offers slightly better performance. Then, the GT variants (explained below) showcase the large potential performance gains between doing and not doing ensemble selection on top of an already pretrained network with no extra fine-tuning. The 'k:' in the legend represents the peak performance w.r.t the number of candidates.}
    \label{fig:results-ensemble-selection-algorithms}
\end{figure}

Interestingly, this is only possible because our model is test-time adaptable by querying it with different maskings for the same sample as well as using different aggregations. While previous work has focused on learnable aggregation functions beyond the simple average \cite{marcu2023self,pirvu2023multi}, little work has been done around the selection process of candidates. We present our results in Figure \ref{fig:results-ensemble-selection-algorithms}, followed by a thorough interpretation.

\textbf{Simple average:} we validate again the performance boost offered by the simple average aggregation (red line) compared to not doing ensembles at all (K=1), with the results swiftly increasing and then plateauing, with a peak at about K=20-50 candidates, which is consistent with the earlier tables and figures. The Mean IoU score of the average of all 100 candidates in this experiment is 54.98 (the red \& orange lines intersecting at k=100). Furthermore, we observe a reduction in Mean IoU variance as we increase the number of candidates, which is expected.

Our first hypothesis upon seeing these results was to think that good candidates will cluster around an average solution. For this we developed a solution that picks the top-k closest candidates around the mean. While this was a good idea on paper, the best score out of many experiments did not improve satisfactorily upon the simple average. Based on this previous result, our next hypothesis was that it is more likely that there are a few strong and diverse candidates rather than more of them clustered around the mean. We perform the following experiment, which uses the test set ground truth for guidance as an oracle. For this reason, while these numbers are not generalizable across different datasets and tasks, they can show us an upper bound of what's possible with the right selection and aggregation algorithm on top of our PHG-MAE ensembles. These are denoted with (GT) in the caption. We compute the IoU of each of the 100 independent predictions for all 116 entries of Dronescapes-Test, resulting in a $116 \times 100$ matrix, one for each frame. For each frame, we sort the candidates by their individual IoU for each frame (i.e. sort each row of the matrix).

\textbf{Greedy-BestK(GT):} this variant (blue dashed line) uses the best per-frame subset of top-k candidates added one by one based on their sorted top-k positions followed by simple average. For example, if K=20 is the best, we check for K=1, K=2 (using top-2 best individuals)..., K=100 (using all 100 of them), perform the average followed by picking the best K subset (20 in this example). Since each frame has its own unique best k subset, this result is unique, hence why it's a single number.

\textbf{Greedy-TopK(GT):} this variant (orange line) also sorts each individual candidate and adds them one by one for each K entry. However, it does not pick the best one, but rather we report all the K results (1..100), similar to Simple average. Note that each frame also has its own top-k best candidates.

While the Greedy-BestK(GT) method provides a great score of 60.42, remarkably, very good results are obtained by only using a single best candidate (out of 100), with a score of 59.94. The results hold well even when using the top-5 best candidates (selected at each frame), with a score of 59.87 (orange line). These results should be interpreted as an upper bound for the potential of better test-time selection algorithms. Using the Simple average method (random top-k, no selection) yields only a score of 55.06, meaning that there are about 4.5 Mean IoU points to be gained by simply discovering better ways to select the best top-1 candidate out of 100 (59.94) without any model re-training or expensive data acquisition. Selecting the best sub-group can add yet another half a point (60.42). Always finding the top-1 out of 100 may be a bit of a stretch, however, as the Greedy-BestK(GT) line shows, we could get some improvement over the current results even with only finding the 20th best candidate without any ensemble at all. Being able to do this would be in line with the recent research on test-time adaptation. However, as individual results get worse, we see a drastic degradation, thus using ensembles may be beneficial if there is no algorithmic way to figure out the individual best performers given a pool of candidates.

\begin{figure}[h!]
    \centering
    \includegraphics[width=1\linewidth]{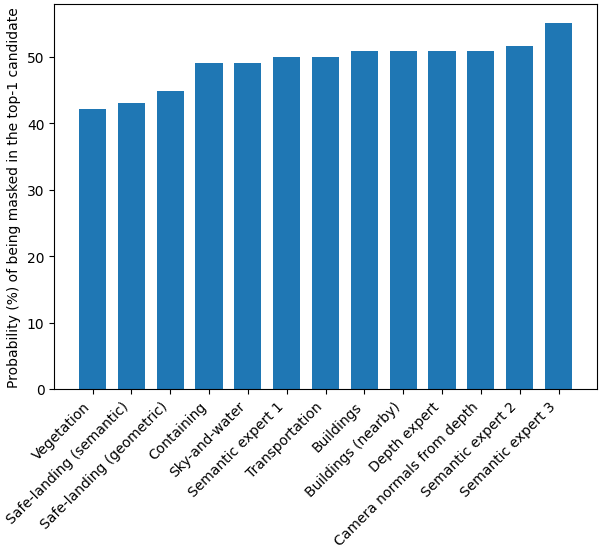}
    \caption{Ensemble candidates selection algorithm: the distribution of the top-1 performing candidates' input modalities across the test set. This chart shows the frequency at which each input modality was masked in the single best-performing prediction (out of 100 candidates, identified using an oracle). Lower bars indicate modalities that were more frequently used (i.e., unmasked) in the top-performing predictions, highlighting their importance. }
    \label{fig:results-top-performer-input-modalities}
\end{figure}

\textbf{Simple average (+masking distribution):}. this variant (green line) in Figure \ref{fig:results-ensemble-selection-algorithms} is obtained by analyzing the distribution of the intermediate modalities with regards to the top-1 candidate. This distribution can be seen in Figure \ref{fig:results-top-performer-input-modalities}. It should be noted that while this method outperforms the simple average one, these weights were derived by analyzing the test set input distribution as an oracle, which may not generalize for other inference data. The purpose of this analysis is to showcase a potential direction for providing cues to the masking algorithm at test time in order to perform better than the standard case.

We observe that the top performing candidate has a tendency to use the intermediate modalities more often than the raw experts. This further justifies the main reason they were introduced and that is to act as a bridge between the low-level RGB data and the high-level output modalities like semantic segmentation (see Section \ref{subsec:intermediate-modalities}). Because of their procedural nature, they are designed to be more generic and thus more robust to domain shifts. For example, the 'safe-landing (semantic)' intermediate modality is an algorithmic combination of semantic segmentation, depth estimation and camera normals derived from depth. We believe that creating even more derived intermediate modalities from experts will further boost the performance of the models as well as provide even more opportunities in the pool of candidates for ensemble candidates selection.

To conclude this section, it should be noted that this line of research is a promising area in the spirit of test-time compute and budget allocation \cite{snell2024scaling,beeching2024scalingtesttimecompute}, which must be further investigated. Our PHG-MAE models are compatible with this paradigm, since the models can be 'queried' or 'prompted' by employing different masking strategies (i.e. optimizing the input distribution) as well as doing ensemble candidates selection algorithms (i.e. searching through output distribution) by using simple heuristics (as described here), learned methods like process reward models (PRMs) or other novel approaches. Moreover, there are also better ways to aggregate the selected candidates, not just using the simple average, such as weighted average or a dynamic-aggregation neural network \cite{pirvu2023multi}.

\subsection{Ablation: Impact of Experts vs. Intermediate Modalities}
\label{subsec:experiments-ablation-study}

\begin{figure}[h]
    \centering
    \includegraphics[width=1\linewidth]{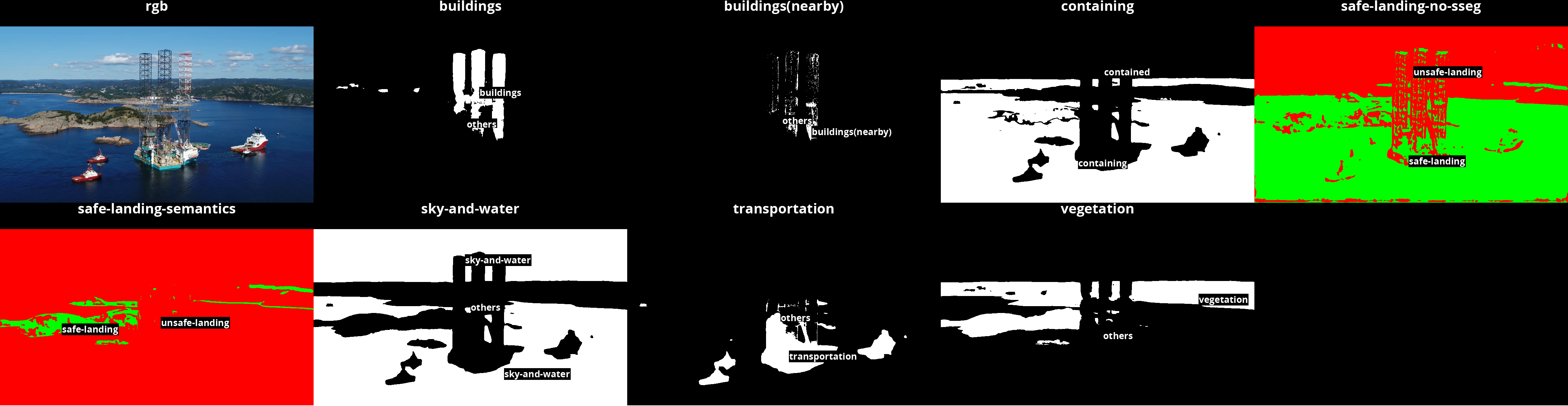}
    \caption{Binary modalities representing the inputs to the PHG-MAE-NRand-B model}
    \label{fig:binary-modalities-sample-qualitative}
\end{figure}

\begin{figure*}[h]
    \begin{minipage}[t]{0.4\linewidth}
      \centering
      \includegraphics[width=1\linewidth]{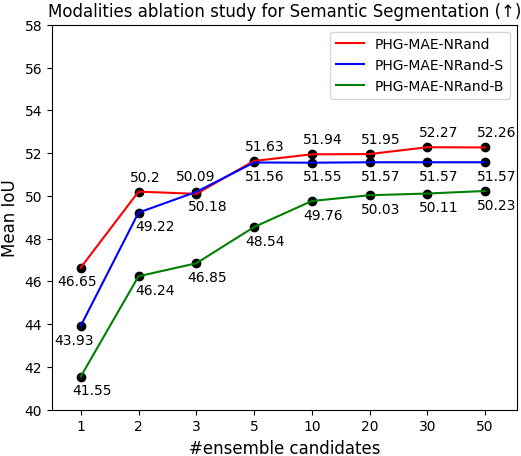}
    \end{minipage}
    \hfill
    \begin{minipage}[t]{0.59\linewidth}
        \vspace{-37mm}
        \centering
        \scalebox{0.775}{
        \begin{tabular}{|c c c c c|}
            \hline
            \vspace{1mm}
            Model & \# Input   & Mean IoU $\uparrow$ & Mean IoU $\uparrow$ & \\ [-1.0ex]
                  & modalities & (no ensembles)      & (ensembles)         & \\ [-0.4ex]
            \hline
            \textbf{PHG-MAE-NRand} & \textbf{14} & \textbf{46.64} & \textbf{52.40 ± 0.22} & \\
            \hline
            \underline{PHG-MAE-NRand-S} & \underline{4} & \underline{43.92} & \underline{51.56} & \\
            \hline
            PHG-MAE-NRand-B & 9 & 41.55 & 50.02 & \\
            \hline
        \end{tabular}
        } 
    \end{minipage}
\caption{Ensembles Modalities Ablation Study. Left: Figure with performance at various ensemble candidate counts. Right: Table of performance with no ensemble and an ensemble of 50. Note: these models are trained on the original 12K samples only (Dronescapes-Semisup1), not the extended one. The -S variant refers to the model trained on the 3 semantic experts only, while the -B variant was trained only on the 8 derived (binary) intermediate representations.}
\label{fig:ensembles-modalities-ablation-study}
\end{figure*}

In this ablation study, we want to analyze the impact of the type of modality (expert or procedurally-derived) with regard to the overall performance on the semantic segmentation task. We train two additional 4.4M-parameters PHG-MAE-NRand models on subsets of modalities on the Dronescapes-Semisup1-M+ dataset (12K samples). The PHG-MAE-NRand-S variant uses only the semantic segmentation experts, namely the three Mask2Former variants, while the PHG-MAE-NRand-B uses only the 8 segmentation-based binary procedurally-generated intermediate modalities. Both models use RGB as input as well. A sample of the NRand-B model's inputs can be seen in Figure \ref{fig:binary-modalities-sample-qualitative}, while the results of this experiment can be seen at Figure \ref{fig:ensembles-modalities-ablation-study}.

We can use the PHG-MAE-1All as baseline reference (Figure \ref{fig:results-sseg-ensembles}), which only scores 39.10 Mean IoU when trained on the 12K samples of Dronescapes-Semisup1, showcasing that in a low-data setup, the additional modalities offer large performance boosts. In terms of ensembling capabilities, the PHG-MAE-NRand-S is only capable of generating $2^3=8$ ensembles, hence why we observe a stagnation in performance as we increase the number of ensembles after about 5 candidates. However, the PHG-MAE-NRand-B model, despite using only binary-derived intermediate modalities like 'sky-and-water' or 'buildings', continues to improve in performance as the number of candidates grows and can generate up to $2^8=256$ ensemble combinations. Finally, we observe that the best overall model is the one that combines both the semantic experts as well as the binary modalities, validating their usefulness. These are indeed the modalities that we used throughout the earlier results as well.

\subsection{Results: State of the Art for Semantic Segmentation}
\label{subsec:experiments-semantic-segmentation}

In the previous subsections, we have focused on our algorithmic and data-level improvements by presenting the PHG-MAE model, the randomly masked MAE ensembles with its test-time performance boost, as well as a study in teacher-student distillation, including in the case of automatic pseudo-labels selection. In this section we summarize all these findings and present our top-4 most relevant models compared to the state of the art on the Dronescapes-Test benchmark. The results can be seen in Table \ref{tb:results-sseg}.

\begin{table*}[h]
\centering
\begin{tabular}{|l c c c|}
    \hline
    Model & Training Dataset            & Parameters & Mean IoU $\uparrow$ \\ [-0.5ex]
    \hline
    \textbf{PHG-MAE-Distil} & \textbf{Dronescapes3-M+Selection}   & \textbf{4.4M} & \textbf{56.27} \\
    \hline
    \underline{PHG-MAE-NRand} & \underline{Dronescapes2-M+}   & \underline{4.4M} & \underline{55.06 ± 0.09} \\
    \hline
    PHG-MAE-NRand & Dronescapes-Semisup1-M+ & 4.4M & 52.40 ± 0.22  \\
    \hline
    PHG-MAE-1All & Dronescapes2     & 4.4M & 52.04 \\
    \hline
    \hline
    Mask2Former \cite{cheng2022masked} & Mapillary \cite{neuhold2017mapillary}           & 216M & 53.97 \\
    \hline
    NHG-LR \cite{marcu2023self} & Dronescapes-Semisup1+2  & 32M & 40.76 \\
    \hline
    SafeUAV-Distil \cite{marcu2023self} & Dronescapes-Semisup1+2-Pseudo  & 1.1M & 40.31 \\
    \hline
    NGC-mean \cite{leordeanu2021semi,marcu2023self} & Dronescapes-Semisup1+2  & 32M & 36.55 \\
    \hline
    SafeUAV-MTL \cite{marcu2018safeuav,marcu2023self} & Dronescapes-Semisup1   & 1.1M & 32.79 \\
    \hline
\end{tabular}
\caption{Semantic segmentation results on the Dronescapes test set. The first four rows denote our models, while the last five ones present the existing work. Our best performing model, PHG-MAE-Distil (row 1), is trained on the filtered pseudo-labels as generated by PHG-MAE-NRand (row 2). We explain the -NRand ensembles in Section \ref{subsec:experiments-ensembles} and the automatic pseudo-labels selection algorithm in Section \ref{subsec:experiments-dataset-selection}. Row 3 refers to our PHG-MAE model trained only on the frames of the original Dronescapes-Semisup1 enhanced with our newly-added modalities. Row 4 refers to our PHG-MAE-1All model trained on the extended scenes, but without any extra modalities, using only RGB as input. First row corresponds to Stage 3, while the next three rows to Stage 2 of Figure \ref{fig:dronescapes-genesis}.}
\label{tb:results-sseg}
\end{table*}

We showcase the effectiveness of our PHG-MAE model as well as our dataset enhancements on two distinct axes: dataset improvements with more samples \& modalities, and algorithmic improvements with MAE ensembles on top of these modalities (Stage 2 presented in Figure \ref{fig:dronescapes-genesis}). Rows 3 and 4 in the table correspond to these two dimensions, leading to the conclusion that both are valid solutions to improve the results compared to the initial Dronescapes work (i.e. rows 6-9). Moreover, when put together, we observe a great improvement leading to a score of 55.06 (row 2). Finally, by adding  yet another set of frames and doing iterative semi-supervised learning via model distillation \& automatic pseudo-labels selection, we obtain a score of 56.27 (row 1), which is our top result for semantic segmentation on the Dronescapes-Test benchmark.  When comparing to the existing state of the art (i.e Mask2Former \cite{cheng2022masked}, row 5), our models yield better results while being almost 2 orders of magnitude smaller in terms of parameters count.

For an apples-to-apples comparison w.r.t the underlying methodology between our methods and the existing work of \cite{marcu2023self} we can make the following parallels: The SafeUAV-Distil entry (row 7, 40.31 mIoU) is trained on the pseudo-labels of NHG-LR (row 6), making it comparable to our PHG-MAE-Distil variant (row 1, 56.27 mIoU). The NHG-LR (row 6, 40.76 mIoU) and NGC-Mean (row 8, 36.55 mIoU) are directly comparable to our PHG-MAE-NRand model (row 2, 55.06 mIoU) which embeds the entire hypergraph into a single model. Finally, the SafeUAV-MTL (row 9, 32.79 mIoU) is directly comparable to the PHG-MAE-1All variant (row 4, 52.04 mIoU).

\subsection{Results: Multi-modal Multi-task Learning}
\label{subsec:experiments-mtl-results}

In this subsection, we are going to compare our method, denoted as PHG-MAE, against our main baselines, namely multi-task CNNs (SafeUAV-MTL) \cite{marcu2018safeuav} (as reported by \cite{marcu2023self}), Neural Graph Consensus (NGC) \cite{leordeanu2021semi}, and the Neural hypergraphs (NHG) \cite{marcu2023self}. We present the results on the Dronescapes test set on the three main tasks: semantic segmentation, depth estimation and camera normals estimation. We present our main results for MTL in Table \ref{tb:results-mtl}. Note that these results for our model are from these checkpoints where the error of each task is minimized.

\begin{table*}[h]
\centering
\begin{tabular}{|c c c c c c|}
    \hline
    Model & Training & Parameters & Semantic $\uparrow$ & Depth $\downarrow$ & Camera Normals \\ [-0.5ex]
          & Dataset  &            & Segmentation        & Estimation         & Estimation $\downarrow$ \\
    \hline
    PHG-MAE-NRand & Dronescapes2-M+ & 4.4M & \textbf{55.06 ± 0.09} & \textbf{16.13 ± 0.03} & \textbf{12.35 ± 0.01} \\
    \hline
    PHG-MAE-NRand & Dronescapes-Semisup1-M+ & 4.4M & \underline{52.40 ± 0.22} & 20.95 ± 0.02 & \underline{12.38 ± 0.01} \\
    \hline
    PHG-MAE-1All & Dronescapes2 & 4.4M & 52.04 & \underline{18.84} & 12.68 \\
    \hline
    PHG-MAE-1All* & Dronescapes-Semisup1 & 1.1M & 39.23 & 19.31 & 13.18 \\
    \hline
    PHG-MAE-1All & Dronescapes-Semisup1 & 4.4M & 39.10 & 20.55 & 13.48 \\
    \hline
    NHG-mean \cite{marcu2023self} & Dronescapes-Semisup1+2 & 32M & 37.58 & 21.81 & 12.40 \\
    \hline
    NGC-mean \cite{leordeanu2021semi} & Dronescapes-Semisup1+2 & 32M & 36.55 & 20.08 & 12.97 \\
    \hline
    SafeUAV-MTL* \cite{marcu2018safeuav} & Dronescapes-Semisup1 & 1.1M & 32.79 & 21.66 & 12.40 \\
    \hline
    \end{tabular}

\caption{Multi-task learning comparison for semantic segmentation, depth estimation and camera normals estimation. Bold represents the best result, while the second best one is underlined. The PHG-MAE-1Rand models sample a random edge (from the distribution of all edges in the hypergraph) every time. For a consistent result (as different edges have different performance), we report the average error and standard deviation across 100 independent runs on the test set. For NGC and NHG, \textit{mean} refers to the ensemble aggregation function.\\
*PHG-MAE-1All is equivalent to SafeUAV-MTL (same model, same training data); however our independent re-implementation improves these results with no dataset or neural network architecture changes. We provide both results for a fair comparison with our more advanced PHG-MAE method, which builds on top of this non masking (i.e. 1All) MTL variant. Our best explanation for this difference for semantic segmentation is the usage of modern learning rate scheduling and data augmentation.}
\label{tb:results-mtl}
\end{table*}

Our main baseline is the SafeUAV-MTL architecture introduced in \cite{marcu2018safeuav} which was also used in the neural-graph related works. It is a UNet-based CNN architecture with dilated convolutions, but no sort of masking or other additional special operations. To compare with this, we train a slightly larger model of 4.4M parameters both on the original dataset as well as on our extended variant. We observe a large increase by simply adding more data, jumping from 39.10 Mean IoU on semantic segmentation all the way to 52.04. As a reminder, the Dronescapes2 variant only uses RGB as input modality, similarly to Dronescapes-Semisup1, but expands the dataset to 80K training data points across 15 UAV scenes, compared to just 12K data points across 7 UAV scenes. See Table \ref{tb:dronescapes-variants} for all the dataset variants that we are using.

Then, we turn our focus on neural-graph based architectures, namely Neural Graph Consensus (NGC) \cite{leordeanu2021semi} and Neural hypergraphs (NHG) \cite{marcu2023self}. While these models are an improvement over the original CNN MTL, the performance boost comes with a large complexity cost in terms of implementation and maintenance. In their works, the edges are explicit and independently trained neural networks, which are manually stitched together to make fixed ensembles.

Finally, we present our models, referred to as PHG-MAE-NRand which embed the neural hypergraph and the inference-time ensembles into a single model. This model is trained using the process outlined in Section \ref{subsec:intermediate-modalities}, which leverages Masked Autoencoders and intermediate modalities derived from experts. This multi-modal MAE approach results in a significant performance boost, with semantic segmentation improving from 39.10 to 46.64 on semantic segmentation, using the same number of data points but enhanced with the newly extracted modalities from the data-pipeline. Using inference-time ensembles the results improve drastically for all three tasks.

It is also worth noting that we train the regression tasks (depth and camera normals estimation) using the simple L2 loss and semantic segmentation with cross-entropy loss, without task-specific adjustments. Incorporating advanced loss functions, such as photometric loss between consecutive frames \cite{zhou2017unsupervised} or plane-fitting consistency between depth and camera normals, could further enhance task performance and alignment.

\begin{figure*}[h]
    \centering
    \includegraphics[width=1\linewidth]{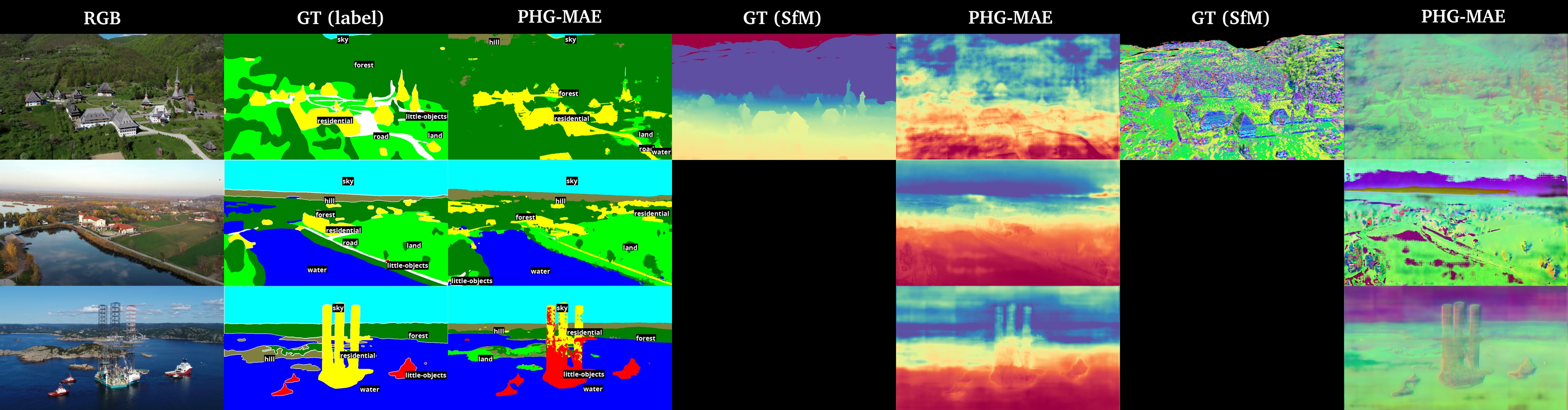}
    \caption{Multi-task learning (MTL) qualitative results from our best model from Table \ref{tb:results-mtl}. The Semantic Segmentation ground truth labels are human-annotated, while for Depth and Camera Normals, it is based on a Structure from Motion reconstruction. Only one scene is available on the original Dronescapes test set for these two tasks.}
    \label{fig:results-mtl-qualitative}
\end{figure*}

\begin{figure*}[h]
    \centering
    \includegraphics[width=1\linewidth]{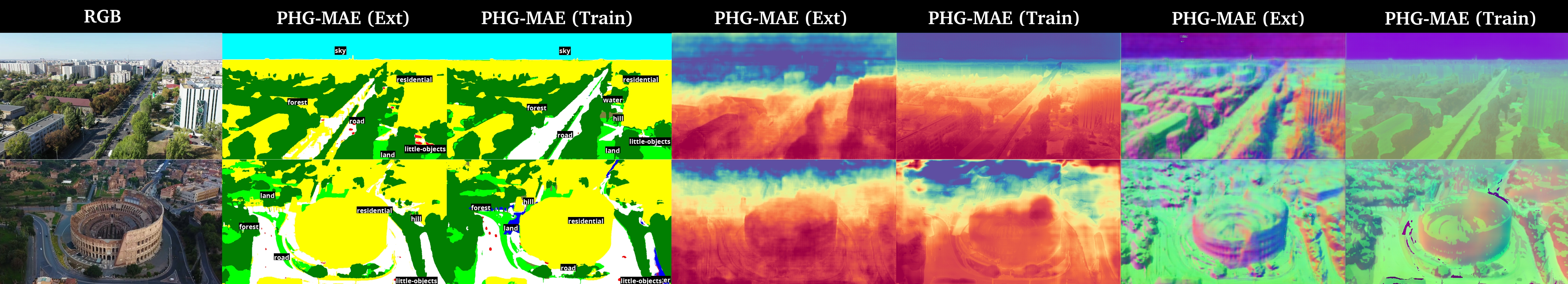}
    \caption{Multi-task qualitative results on unseen videos. PHG-MAE (Ext) refers to our model trained on Dronescapes2-M+, while PHG-MAE (Train) to the model trained on Dronescapes-Semisup1-M+.}
    \label{fig:results-mtl-qualitative-new-scene}
\end{figure*}

\textbf{Qualitative Results} In Figures \ref{fig:results-mtl-qualitative} and \ref{fig:results-mtl-qualitative-new-scene}, we present qualitative samples to contextualize the quantitative results presented earlier.

In the first figure, we show samples from the Dronescapes-Test dataset, with black images where there is no available ground truth due to lack of a SfM-based reconstruction in the original work. Nonetheless, our model still manages to capture relevant information on these tasks as well. For semantic segmentation, the results are similarly consistent, but the inherent ambiguity of the task makes alignment with human annotators challenging. For instance, our network predicts the bottom half of the oil rig (third row) as 'little-object' (a class including boats), whereas annotators labeled it as 'residential' (building).

The second figure presents qualitative results from two completely unseen video scenes. It should be noted that the scene in the first row was included in Dronescapes2-M+ (but not that particular video), showcasing how performance on a specific scene can be drastically improved via fine-tuning if the deployment environment is known in advance, which is a valid use-case in practice (i.e. agriculture or building surveillance). One only needs to fly an UAV around that scene a few times and run our data-pipeline, followed by fine-tuning and potentially knowledge distillation. The second row is from a completely unrelated video downloaded from the internet showcasing generalization.

\begin{figure}[h]
    \centering
    \includegraphics[width=1\linewidth]{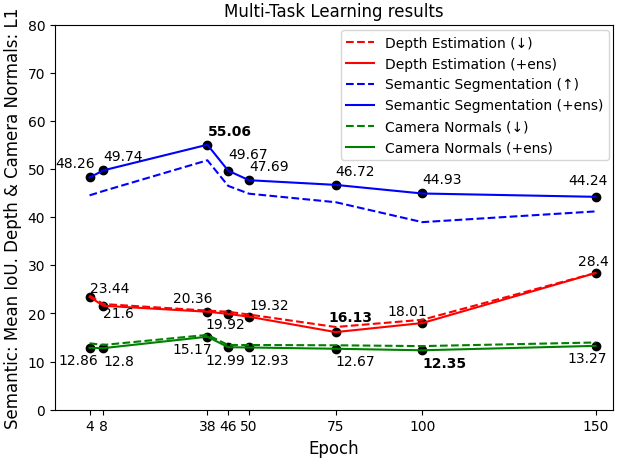}
    \caption{Multi-task learning results per epoch for the PHG-MAE model trained on Dronescapes2-M+. The (+ens) variants use the -NRand variant which uses N=30 random ensemble candidates, while the other variants are the -1Rand ones. The best results of this plot across all three tasks were reported in Table \ref{tb:results-mtl}. This plot highlights the challenge of negative transfer in MTL where the best epoch for semantic segmentation (i.e. 38) differs from the best epochs of depth estimation (i.e. 75) or camera normals estimation (i.e. 100), meaning that the best shared checkpoint is a compromise between each task's performance.}
    \label{fig:results-mtl-per-epoch}
\end{figure}

\textbf{Discussion about Multi-Task Learning.} In Figure \ref{fig:results-mtl-per-epoch} we present the per-epoch results of our PHG-MAE model across all three tasks.

We can observe the difficulty of managing three tasks with a single model: the peak performance of one task happens to the detriment of the other two. This is called \textit{negative transfer} in the MTL literature and happens because each task has its own minima and, as the tasks are not completely correlated, they do not share a common minimum.

An interesting observation is that the best result for semantic segmentation happens early on, at epoch 38, while the results of the depth estimation and camera normals happen later on around epochs 75 or 100. These two latter tasks are clearly more correlated to each other (i.e. both are based on the scene's underlying geometry and physics), as their curves point toward a more shared local minima, while the semantic segmentation curve continuously goes down.

Moreover, our training loss is also not tailored for this type of learning: we are simply averaging the loss of the three tasks. A simple improvement to this would be to dynamically weight up or down the loss based on the running magnitude of the error during training \cite{liu2019loss}. We aim to explore better suited multi-task learning strategies as well as better tasks-specific loss functions, however, for now, it's better for us to use three different checkpoints, each optimized for its own task to maximize the results at the cost of inference time and memory usage.

\section{Conclusions and Future Work}

In conclusion, we introduce a new semi-supervised multi-modal MAE-based model which enables test-time ensembling, called PHG-MAE, through the lens of Probabilistic hypergraphs. PHG-MAE can be applied on top of any Masked Autoencoders model (e.g. CNNs, transformers), small or large. This method aims at simulating a neural hypergraph for multi-modal multi-task learning through random masking modalities, integrated into a single unified PHG-MAE model. We tested our algorithm using small CNN-based models (suitable for UAV tasks that require real-time performance), having between 150K and 4.4M parameters, on the Dronescapes-Test benchmark with models trained on commodity hardware in a MTL context.

For training PHG-MAE, we introduce an open-source data pipeline that enables these ensembles through the use of intermediate modalities extracted from pretrained experts plus procedurally generated ones. Our approach is general, as it can be further used on any arbitrary video to extend and diversify the dataset. We show that the ensembles outperform the classical MTL prediction paradigm by producing higher quality and more temporally consistent semantic segmentation maps even with the simple average ensembling method. Finally, we show that efficient distillation can be applied on top of this complex process to enable real-time inference on commodity hardware, with minimal performance degradation, outperforming much larger models with two to three orders of magnitude more parameters.

Based on our contributions and findings presented, we propose appropriate next steps and future directions:

First, we included intermediate modalities as a bridge between low-level RGB and high-level semantic segmentation such that the model can learn a smoother transformation. We aimed for modalities that are more general, rather than dataset-specific, and thus more robust to distribution changes. A future direction can be to formalize the level of abstraction between low (i.e. RGB), mid (i.e. depth estimation) and high (i.e. semantic segmentation) so we can quantify whether more abstract or more concrete modalities boost the learning performance. Furthermore, while our data-pipeline supports automatic generation on new videos, these modalities are still hand-crafted by us. Instead, they could be discovered automatically by designing suitable self-supervised learning metrics.

Second, while our data-pipeline allows to create datasets in an automatic fashion, we did not analyze what type of videos are required for efficient dataset creation, nor how many of them. We added 16 more unlabeled videos (in two separate iterations), which boosted significantly the performance through the unsupervised learning process. However, we do not know how much more can be gained in performance by a potentially continuous addition of videos. The process of adding new data can also be guided by failure or low confidence cases from production environments, such as \cite{tesla-data-engine}.

Third, this approach can also be extended to other domains where video data is available, such as indoor robotics, autonomous driving, action learning from videos and so on. Fourth, while our research was focused on efficient CNNs with few parameters (150k $\sim$ 4.4M), it would be worthwhile to test our methods on larger models with different architectures as well (i.e. attention-based transformers). Fifth, while we used a task-specific loss for semantic segmentation, for the tasks of depth and camera normals estimation we used the simple L2 loss. Using a more task-aligned loss could further improve the results. Also, adding more than just three tasks would be interesting to test. Sixth, we performed a simple average as an aggregation function, however, as we showed, using a better selection algorithm (out of many candidates) can yield up to 5 points in semantic segmentation performance without any training. Furthermore, we only used a fixed masking algorithm where each intermediate modality is masked with a 50\% probability at each inference step resulting in a sampled hyperedge. Further research is needed on exploring better masking strategies, selection and ensemble aggregation functions. Finally, while our distilled models were tested in an experimental setting, actually deploying these models on real UAVs or robots could have a great and direct impact on improving learning efficiency and robustness in the real world.

\textbf{Acknowledgements.} This work is supported in part by projects “Romanian Hub for Artificial Intelligence - HRIA”, Smart Growth, Digitization and Financial Instruments Program, 2021-2027 (MySMIS no. 334906) and "European Lighthouse of AI for Sustainability - ELIAS", Horizon Europe program (Grant No. 101120237).

\appendix

\section{PHG-MAE Model Additional Details}

\subsection{Modeling Multi-Modal Neural Hypergraphs with a Single Neural Network}
\label{subsec:modeling-neural-hypergraphs}
In this section, we show a visual representation of using a single CNN is equivalent to using a multi-modal neural graph. We start with the following figure.

\begin{figure}[h]
    \centering
    \includegraphics[width=1\linewidth]{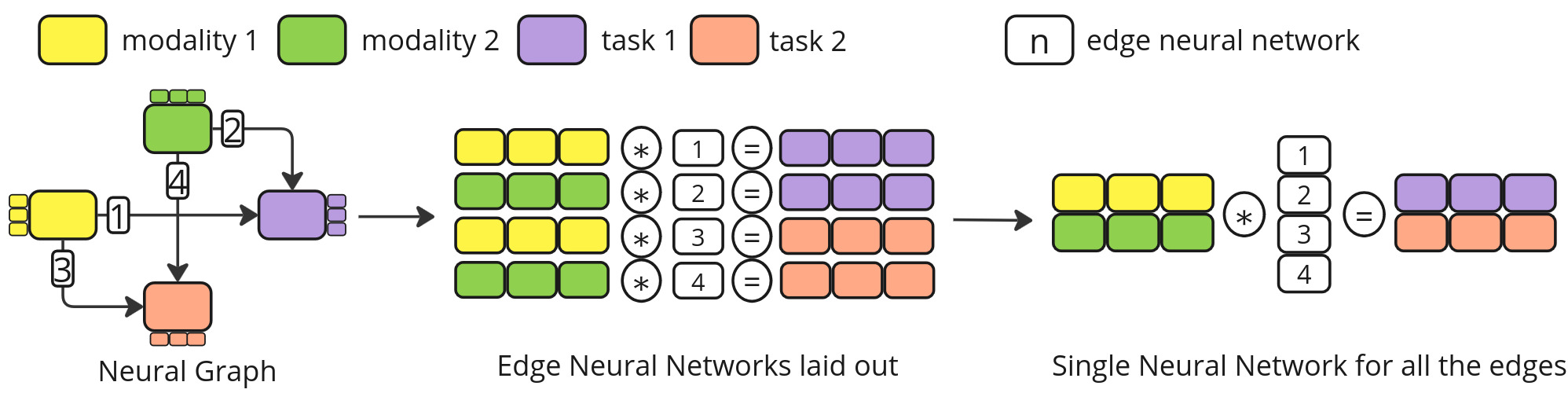}
    \caption{Converting a neural graph (left) of 2 input \& 2 output nodes to four independent edge neural networks (middle) followed by fusing the four networks into a single neural network. We try to prove that this fusing is possible and under what conditions.}
\end{figure}

What we'd like is to have a formulation such that the form in the middle is somehow equivalent to the one on the right. This would be possible if the white boxes on the right (1, 2, 3, 4) can be modeled with a single neural network. Assuming that this equivalence is possible, then all we'd need to do is to properly mask the inputs and the weights to reconstruct all the four edges, like in the figure below. Visually, what we try to achieve is in the next figure. We present a proof below showcasing that the above proposition holds for convolutional neural networks. We start with a simple example of two independent convolutional layers performing 1D convolution on two independent inputs (yellow and green). In the figure below, we show that through proper masking either at input level, or at convolutional filters level, we can recover both of the initial layers. In this case the inputs are 1D vectors of shape $(5,)$ convolved by a filter of shape $(2,)$ resulting in 1D vectors of shape ($4,)$.

\begin{figure}[h]
    \centering
    \includegraphics[width=1\linewidth]{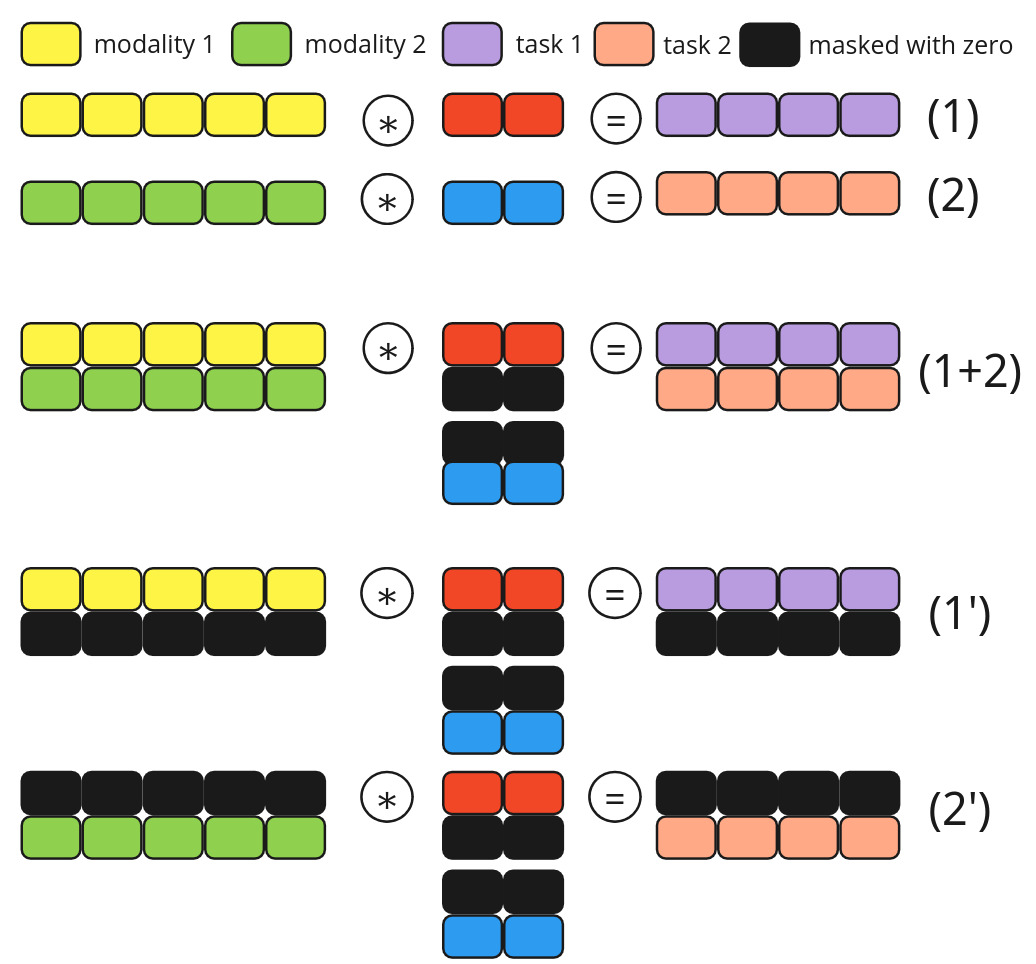}
    \caption{Fusing two independent convolutions (1) and (2) into a single one by masking. The row denoted as (1+2) shows how both outputs can be recovered by adding one extra dimension to the convolution (2D) and masking the correct slice in each dimension. The last two rows (1') and (2') show how by further masking the right input slice on top of (1+2), we can recover each individual 1D convolution showcasing the equivalence.}
    \label{fig:simple_example_conv_singlenet}
\end{figure}

In the top two rows of the figure (labeled as 1 and 2), the two input vectors are independently convolved by a single 1D kernel each (red and blue) resulting in the two individual output vectors. Then, in the next row (labeled as 1+2), we see that by stacking the two inputs together and adding a set of zeroed out weights (one for each filter), we can reconstruct the two 1D convolutions with a single 2D convolution. Furthermore, on the last two rows (labeled as 1' and 2'), we see the two original convolutional layers recreated through masking both the inputs and the filters on top of the bottom left stacking. This is the basis for showing the equivalence between NGC with single-hop Edges (E) and a single neural network model with proper input and weight masking. But, we can go even further with it:

\begin{figure}[h]
    \centering
    \includegraphics[width=1\linewidth]{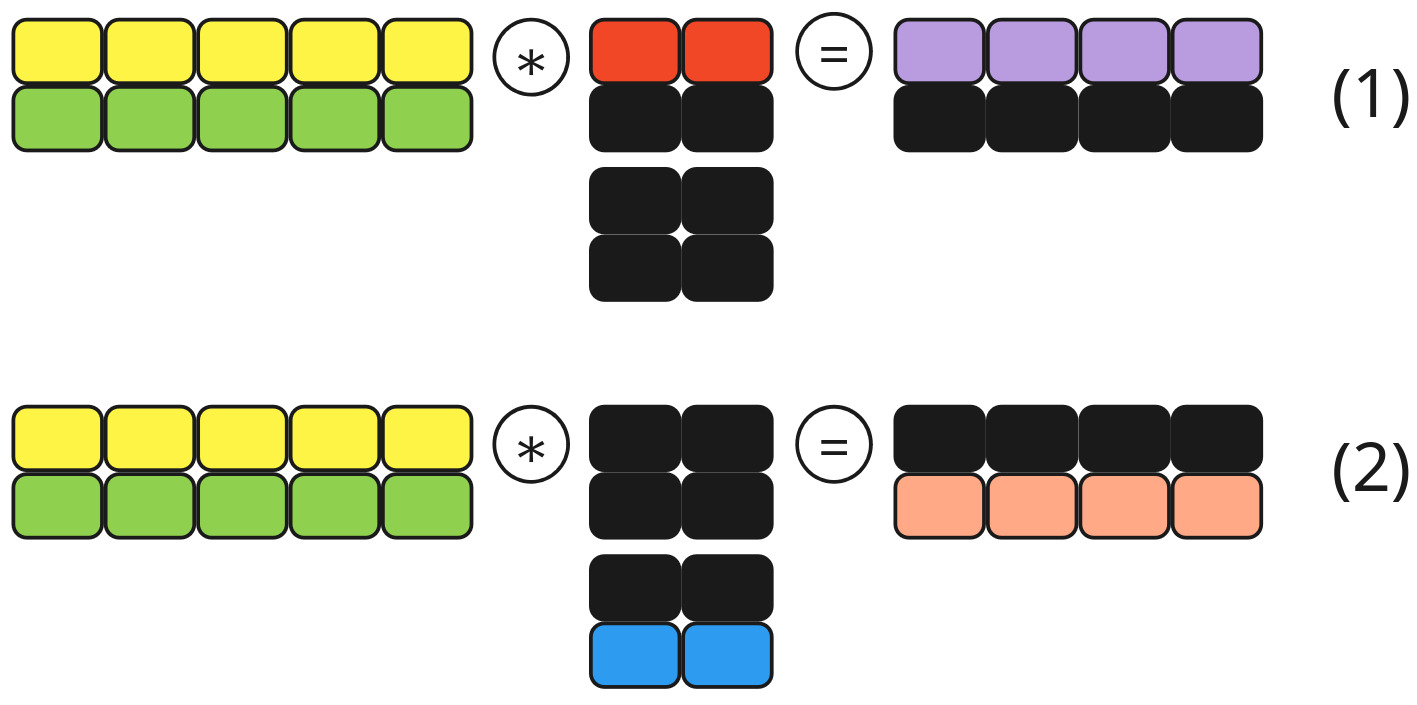}
    \caption{Two Aggregation Hyperedges (AH) from Neural hypergraphs \cite{marcu2023self}. The same kernels are used as in Figure \ref{fig:simple_example_conv_singlenet}, showing that the same network supports both edges and hyperedges with proper masking operations.}
\end{figure}

In the figure above, we are modeling two Aggregation Hyperedges (AH) with the same unique 2D convolution by masking the filter weights. \\

\begin{figure}[h]
    \centering
    \includegraphics[width=1\linewidth]{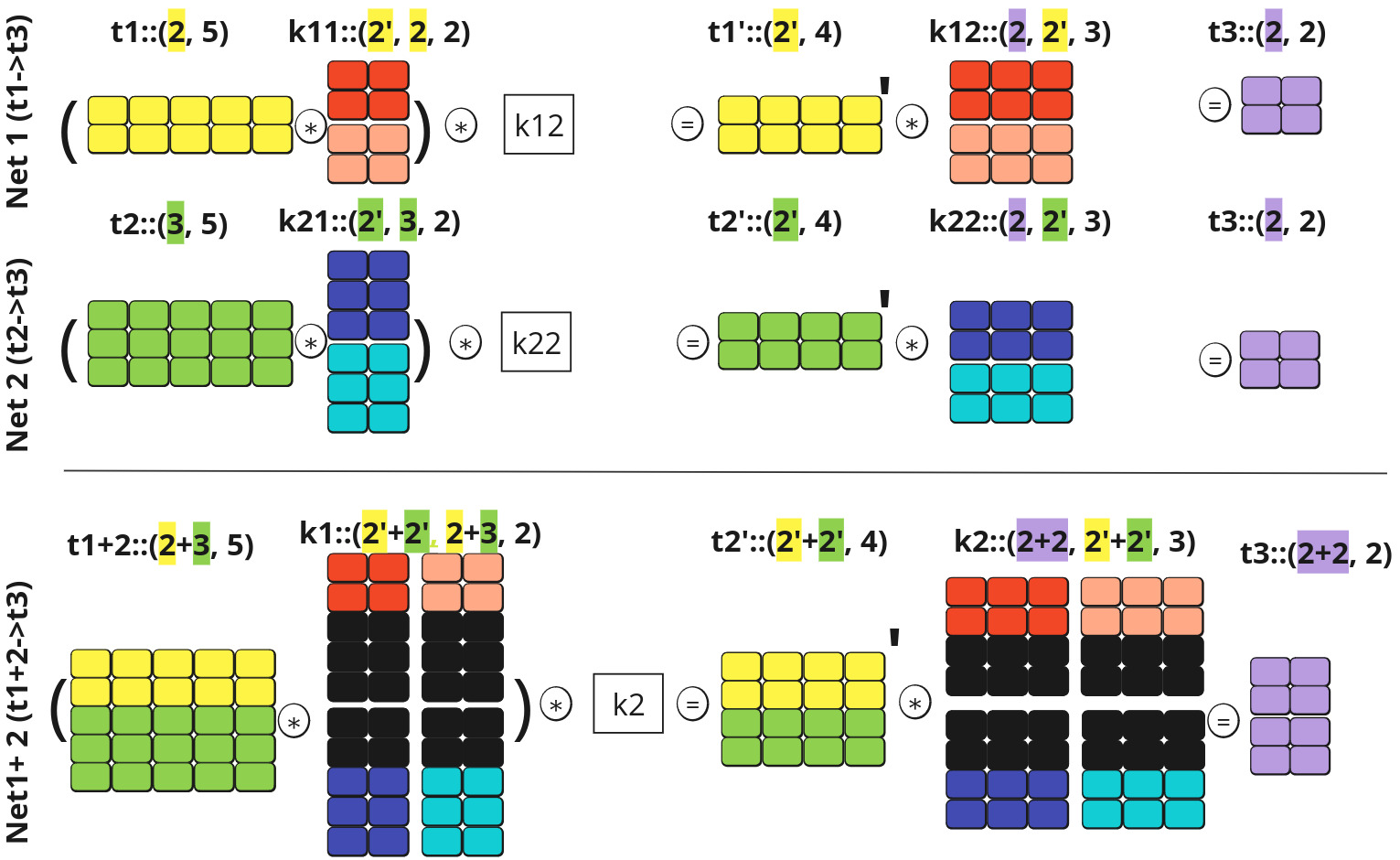}
    \caption{Two-layer network generalization. The two layers can be decomposed into two single-layer operations. Furthermore, the inputs are now two dimensional vectors, showcasing the generality and equivalence of this masking strategy from neural hypergraphs to a single fused network.}
    \label{fig:two_layer_example_convnet}
\end{figure}

\textbf{Two layers and multiple modality channels:}
in Figure \ref{fig:two_layer_example_convnet} we show a more complex example, that is closer to the experiment we run. It turns out that we can apply this type of masking to deep neural networks as it factors out into a layer-by-layer operation, reducing to the case above. For multi-dimensional inputs, such as RGB images we can apply vector linearization (e.g., reshaping $(C,H,W)$ inputs to $(C,H \times W)$), followed by applying the same operations. The same can be applied to multi-dimensional output modalities. Next, in Figure \ref{fig:two_layer_example_convnet}, we present the same steps as above, but with multi-dimensional inputs and a two-layered convolutional neural network. As the convolution is a special case of a fully connected layer, the same strategy applies to fully connected layers.

In this example we also have two input modalities (yellow and green) and a single output modality (violet). This time the number of channels of the modalities are: two (yellow), three (green) and two (violet) while the length (equivalent of image resolution) is 5. Different length modalities would require extra padding or interpolation, similarly to the case of different resolution images on the 2D case. The example showcases applying two convolutions in a row, the equivalent of a two-layer neural network. Starting with the top left convolution: the yellow modality has two associated kernels for the first layer (red and pink) and results in the yellow' feature map that also has two channels. Since the intermediate feature map yellow' has two channels as well, we have two kernels (red and pink). Each kernel (red and pink) has two rows because the yellow modality has 2 channels. The same can be seen for the green modality and the blue kernel which has 3 rows. Then the yellow' feature map is convolved with another set of kernels (red' and pink') of length 3, resulting in the final violet output, that is modality 3 with 2 channels and length 2.

All the operations presented here are just generalizations of the same logic presented in Section \ref{subsec:modeling-neural-hypergraphs}, with the addition of extra layers (one intermediate feature map) and extra channels for each modality (extra kernels and extra rows for each kernel). Following the same logic, we zero out for the red and pink kernels the added channels of the combined input (yellow + green) and we do the opposite in the bottom two kernels, yellow and cyan, where we zero out the channels associated with the red and pink channels.

One important note: the number of rows of zeros that are added for each modality's kernels is equal to the sum of all the channels of all the other modalities, thus naively implementing this process requires quadratically more memory and computation in the number of modalities compared to a simple convolution operation. Specialized algorithms that take into account this sparsity could, in theory, turn the operations of the merged network into N separate networks, basically reverting the merging during the computation, leading to the original complexity that scales linearly with the number of modalities. \\

\textbf{Beyond CNNs:} We also found that attention-based operators generalize similarly by modifying the attention matrix, however, in this work we focus on CNNs only, leaving room for new architectures as further research.

Furthermore, we can even train a single neural network to embed multiple ones using the same technique. By initializing these kernel inputs with zeros, gradient-based optimization effectively ignores them, as their contributions remain zero. One downside of this technique is the requirement of zeroing out the weights for each network that we want to embed, making the memory requirement grows quadratically with the number of networks. Specialized implementations that take care of this sparsity at weights level may allow for further optimizations. Moreover, we only focused on single-step edges (E and AH). For two-hop edges (TH, EH and CH), we need a two-step operation as well (i.e. the outputs are convolved again by potentially other masking). These edge types are not covered in this work.

In conclusion, we show an equivalence between the neural graphs and hypergraphs presented in \cite{leordeanu2021semi,marcu2023self,pirvu2023multi} with a single neural network and proper masking at inputs and weights level. In the next section we move to see how we extend Masked Autoencoders (MAE) for multi-modal multi-task problems, with a focus on, but not limited to, aerial image understanding on Dronescapes.

\subsection{Training Times}

This section presents in Table \ref{tb:training-times} the training durations of various models and datasets used in this work.

\begin{table*}[h]
\centering
\scalebox{0.985}{
\begin{tabular}{|c c c c c|}
    \hline
    Model & Parameters & Dataset & Dataset Size     & Training Duration \\ [-0.5ex]
          &            &         & (I/O modalities) & \\
    \hline
    SafeUAV-MTL & 1.1M & Dronescapes-Pseudo1 & 12K (1/3) & 7.4h \\
    \hline
    SafeUAV-MTL & 4.4M & Dronescapes2-M+ & 80K (1/3) & 49.3h \\
    \hline
    PHG-MAE & 4.4M & Dronescapes2-Train-M & 12K (14/3) & 21.9h \\
    \hline
    PHG-MAE & 4.4M & Dronescapes2-M+ & 80K (14/3) & 145.9h \\
    \hline
    PHG-MAE-Distil & 1.1M & Dronescapes3-M+Pseudo & 148K (1/1) & 35.47h \\
    \hline
    PHG-MAE-Distil & 4.4M & Dronescapes3-M+Pseudo & 148K (1/1) & 70.13h \\
    \hline
\end{tabular}
} 
\caption{Training duration of various models and datasets used in experiments. All the times are based on training them on a server with 8xRTX2080 Ti GPUs for 150 epochs. Details about the performance of each model are presented in the next sections. The I/O modalities column represents the number of inputs and outputs. For example in PHG-MAE, 14 modalities are input and 3 are output.}
\label{tb:training-times}
\end{table*}

\subsection{Analysis of per-scene similarity distribution for unsupervised distillation on Dronescapes3-M+Pseudo}

In this section we provide a more comprehensive view of the data distribution and thresholds of each scene used to perform automatic pseudo-labels selection for unsupervised distillation. In Table \ref{tb:similarities-per-scene}, we provide the number of frames before and after applying the global threshold based on the 20th percentile of similarities. The global thresholds for 20th and 25th percentiles (used in experiments) were 0.835 and 0.813.

\begin{table}
\scalebox{0.79}{
\begin{tabular}{lllll}
\toprule
scene & Count & Mean       & Count    & Used(\%) \\
      &       & Similarity & Filtered & after filter \\
\midrule
ovaselu\_DJI\_0..\_540p & 9022 & 0.72 & 0 & 0.00 \\
herculane\_DJI..\_540p & 8204 & 0.66 & 12 & 0.15 \\
jupiter\_DJI\_0..21715 & 1331 & 0.60 & 11 & 0.83 \\
olanesti\_DJI\_..\_540p & 9765 & 0.68 & 112 & 1.15 \\
politehnica\_D..\_540p & 9021 & 0.74 & 139 & 1.54 \\
castelulcorvi..\_540p & 3121 & 0.74 & 56 & 1.79 \\
paris\_youtube\_1\_540p & 8455 & 0.63 & 294 & 3.48 \\
sanfrancisco\_..\_540p & 5635 & 0.57 & 233 & 4.13 \\
norway\_DJI\_0708\_540p & 4763 & 0.66 & 475 & 9.97 \\
riodejaneiro\_..\_540p & 6264 & 0.57 & 692 & 11.05 \\
sheerness\_you..\_540p & 4950 & 0.76 & 759 & 15.33 \\
herculane\_DJI..\_full & 726 & 0.77 & 122 & 16.80 \\
comana\_DJI\_20..\_540p & 9315 & 0.74 & 1782 & 19.13 \\
olanesti\_DJI\_..\_full & 1089 & 0.80 & 235 & 21.58 \\
slanic\_DJI\_09..\_9780 & 9001 & 0.80 & 1954 & 21.71 \\
horezeu\_DJI\_2..\_540p & 8992 & 0.75 & 2024 & 22.51 \\
slanic\_DJI\_20..\_540p & 8985 & 0.79 & 2657 & 29.57 \\
raciu\_DJI\_0418\_540p & 9022 & 0.83 & 4395 & 48.71 \\
gradistei\_DJI..13110 & 1089 & 0.85 & 678 & 62.26 \\
petrova\_DJI\_0..11850 & 726 & 0.86 & 468 & 64.46 \\
pietrosa\_DJI\_..\_540p & 9762 & 0.83 & 6998 & 71.69 \\
atanasie\_DJI\_..\_full & 4689 & 0.86 & 3659 & 78.03 \\
\bottomrule
\end{tabular}
} 
\label{tb:similarities-per-scene}
\caption{Dronescapes3-M+Pseudo scene distribution before and after applying the global threshold of the 20th percentile of similarities.}
\end{table}

\begin{figure}[ht!]
    \centering
    \includegraphics[width=1\linewidth]{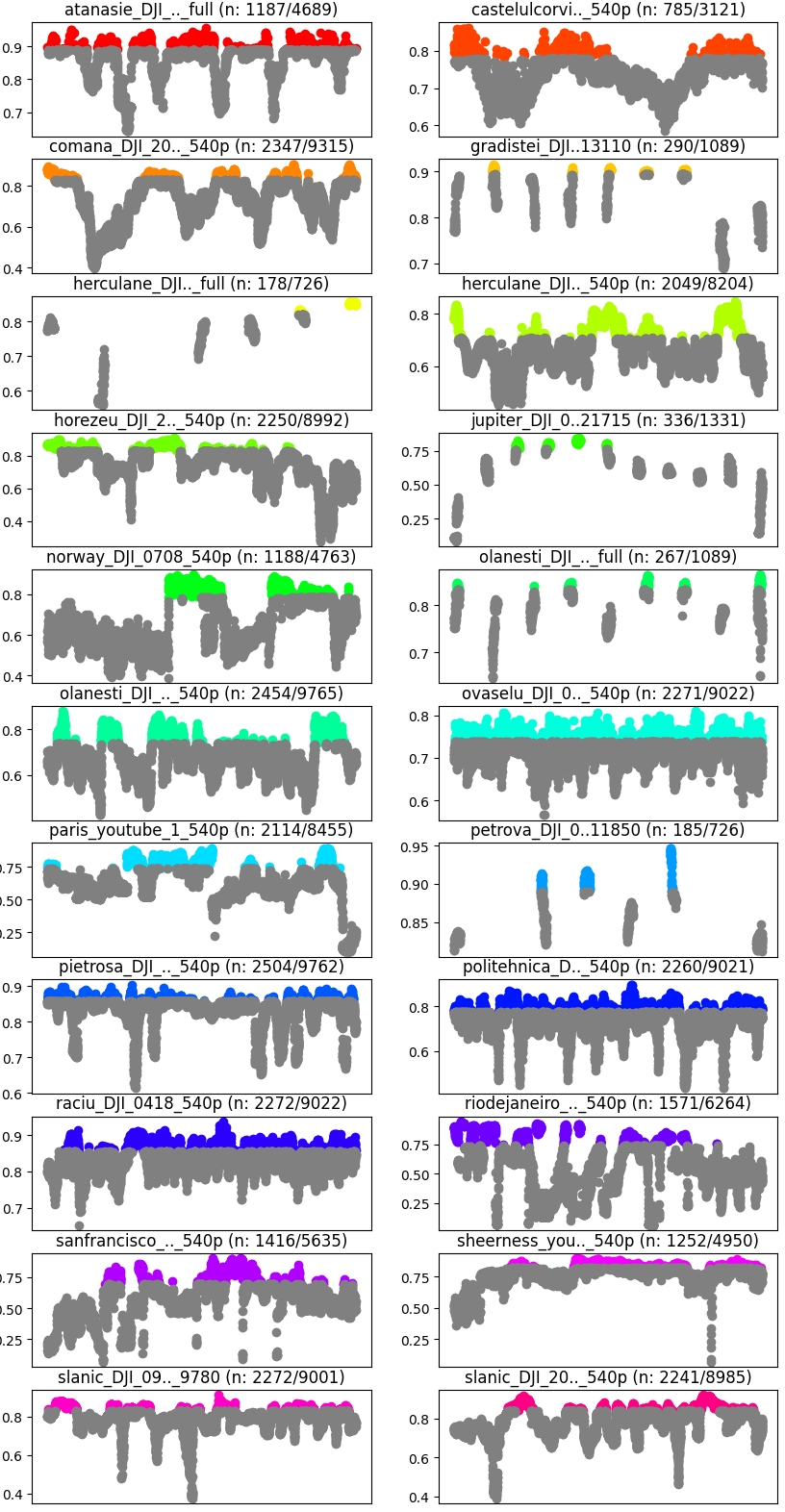}
    \caption{Automatic pseudo-labels selection. The per-scene kept and discarded pseudo-labels based on the average similarity of the candidates to the ensemble. We kept the top 25\% most consistent pseudo-labels.}
    \label{fig:dataset-selection-per-scene-thresholds}
\end{figure}

As some scenes would contain none or a few dozen frames only, this would lead to a lower representation and diversity in the pseudo-labels. As we've seen, this kind of selection provided worse results on the distillation. As a consequence, we performed a per-scene threshold and picked the top 20th and 25th pseudo-labels of each scene, rather than using a global threshold. The selected pseudo-labels for each scene (top 25\%) can be seen in Figure \ref{fig:dataset-selection-per-scene-thresholds}

Interestingly, each scene also has its own sub-distribution that sometimes follows a cyclic path. This corresponds to the actual videos which cycle through the same scene while flying. It is possible that some regions are simpler to predict, thus more consistent across candidates and pseudo-labels.

\subsection{Analysis of top-1 performer per scene for the entire test set}

\begin{figure}[ht!]
    \centering
    \includegraphics[width=1\linewidth]{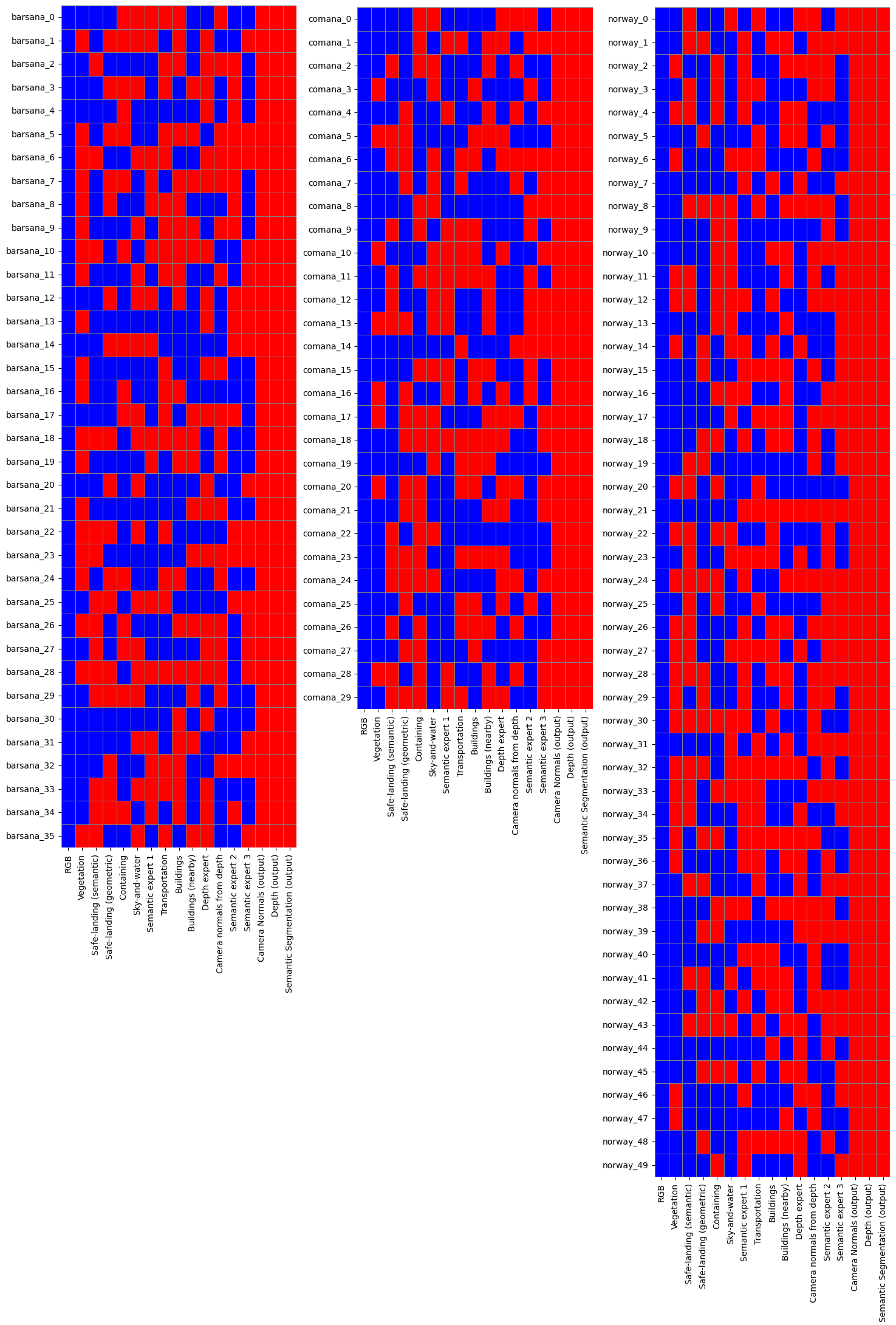}
    \caption{Top-1 performer of each test set entry out of 100 candidates. Blue means not masked, red means masked.}
    \label{fig:results-top-performer-full-test-set}
\end{figure}

In Figure \ref{fig:results-top-performer-full-test-set}, we analyze the entire test set w.r.t the masking distribution.

These are the results that were used to produce the masking distribution presented in Figure \ref{fig:results-top-performer-input-modalities}. Interestingly enough, for each frame there is a unique masking distribution and there doesn't seem to be a strong pattern even for the same scenes and a slight correlation for nearby frames. This means that finding the top-1 performer requires an adaptive algorithm based on multiple candidates.

\subsection{Iterative semi-supervised distillation vs. regular distillation}

In this subsection we want to provide our hints into introducing new data vs. not doing so. We already showed in the main paper that doing iterative semi-supervised distillation provides a great benefit. We maintain a steady of score from 55.06 (teacher) to 55.05 (student) when adding 58K more frames on our 4.4M model. Moreover, we improve the results from 55.06 all the way to 56.27 with dataset pseudo-labels selection as well. But what if we don't have new data? We present the results in Table \ref{tb:distillation-without-incremental}.

\begin{table*}[h]
\centering
\scalebox{0.94} {
\begin{tabular}{|c c l c|}
    \hline
    Model & Parameters & Training   & Mean IoU $\uparrow$ \\
          &            & Dataset    &                     \\
    \hline
    \textbf{PHG-MAE-NRand} & \textbf{4.4M} & \textbf{Dronescapes2-M+ (80K)} & \textbf{55.06 ± 0.09} \\
    \hline
    \underline{PHG-MAE-Distil} & \underline{4.4M} & \underline{Dronescapes3-M+Pseudo (148K) } & \underline{55.05} \\
    \hline
    PHG-MAE-Distil & 430k & Dronescapes3-M+Pseudo (148K) & 54.94 \\
    \hline
    PHG-MAE-Distil & 1.1M & Dronescapes3-M+Pseudo (148K) & 54.30 \\
    \hline
    PHG-MAE-Distil & 4.4M & Dronescapes2-M+Pseudo (80K) & 54.27 \\
    \hline
    PHG-MAE-Distil & 150k & Dronescapes3-M+Pseudo (148K) & 53.32 \\
    \hline
    PHG-MAE-Distil & 430k & Dronescapes2-M+Pseudo (80K) & 52.44 \\
    \hline
    PHG-MAE-Distil & 1.1M & Dronescapes2-M+Pseudo (80K) & 51.80 \\
    \hline
    PHG-MAE-Distil & 150k & Dronescapes2-M+Pseudo (80K) & 50.90 \\
    \hline
\end{tabular}
} 
\label{tb:distillation-without-incremental}
\end{table*}

The distillation results seem to only hold when adding new data. However, upon doing pseudo-labels selection using the feedback from the teacher, these results can improve. It would be interesting to do a self-cyclical training where the feedback from the teacher can be used to train more efficient distillation without any extra data or even re-training the PHG-MAE teacher model itself on the reduced frames.

\section{Dronescapes2 Dataset Additional Details}

\subsection{Data-pipeline: Experts, Intermediate Modalities and Mappings}

The data-pipeline used to compute the extended dataset is an open-source project and can be found at: \url{www.github.com/meehai/video-representations-extractor}.

\begin{figure*}[h]
    \centering
    \includegraphics[width=0.95\linewidth]{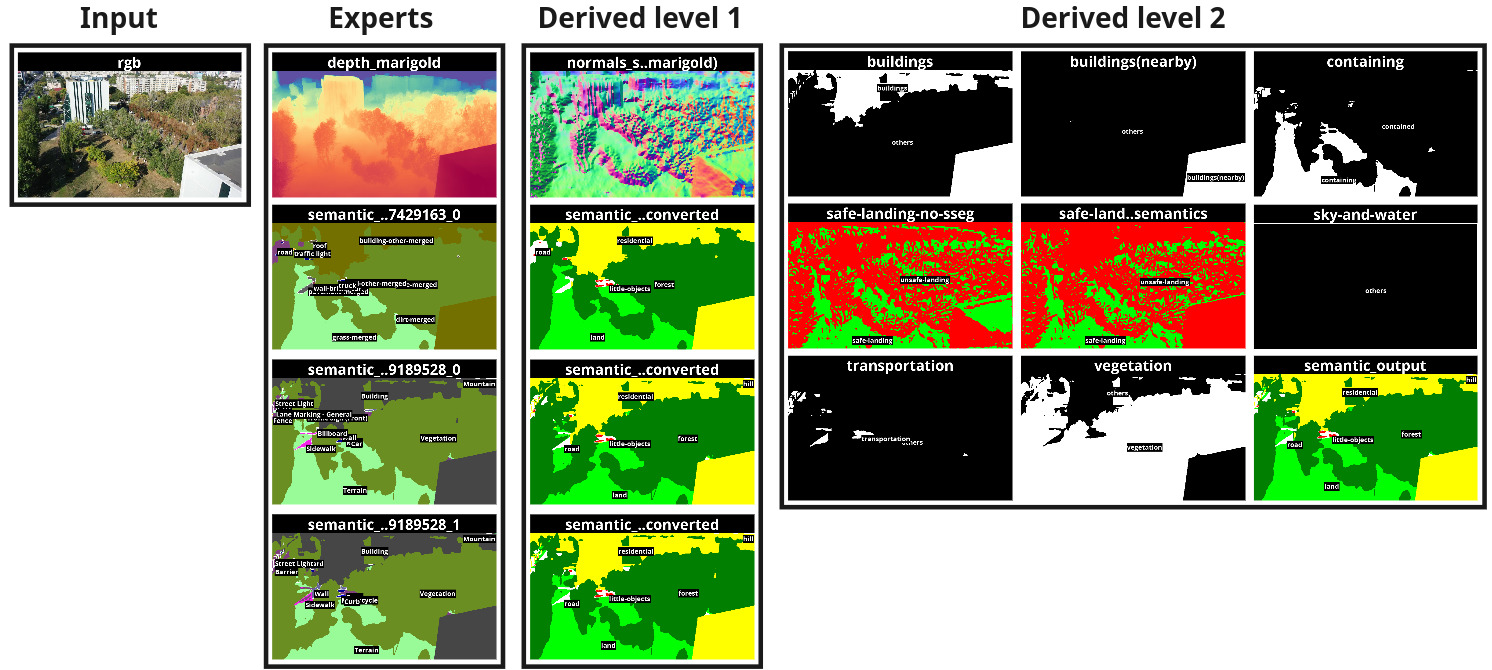}
    \caption{All the extracted experts and derived intermediate modalities in the data-pipeline. Only the mapped Mask2Former variants are used during training (from derived level 1).}
    \label{fig:data-pipeline-full}
\end{figure*}

In Figure \ref{fig:data_pipeline} we used a simplified dependency graph for the intermediate modalities. The full data-pipeline can be seen in \ref{fig:data-pipeline-full}.

As we can see, all the data stems from RGB inputs only, as previously discussed. This enables easy integration with new videos, since all the code depends on this raw information only. The experts are the first level of exported outputs. Both Mapillary (all 3 variants) and Marigold use only RGB as inputs. Then, we have the level-2 derived outputs, which use information only from the experts. Note that the data-pipeline does a topological sort and items from any level $n$ can use inputs from any levels $1 .. n-1$, so in this case, the derived data can also use RGB as input, though it's not needed here. The derived experts here are the camera normals using the SVD algorithm directly from the Marigold expert and an intermediate semantic segmentation that maps the COCO and Mapillary classes to the 8 Dronescapes classes (see the mappings below). Then, the level-2 derived are the intermediate modalities that enable Masked Autoencoding ensembles. Finally, we have a median map of all three Mask2Former outputs, which acts as ground truth for the new videos of Dronescapes2-M+, where no human-annotated data exists, as is the case with Dronescapes-Semisup1.

In our largest Masked Autoencoder Model, we use the following modalities as inputs and outputs:
\begin{itemize}
    \item input: RGB, Buildings, Buildings (nearby), Containing, Camera normals from depth [normals-svd(depth-marigold)], Safe landing (geometric) [safe-landing-no-sseg], Safe landing (semantic) [safe-landing-semantics], Vegetation, Sky-and-water, Transportation, Depth expert [depth-marigold], Semantic expert 1 [semantic-mask2former-swin-mapillary-converted], Semantic expert 2 [semantic-mask2former-swin-coco-converted], Semantic expert 3 [semantic-mask2former-r50-mapillary-converted]
    \item output: depth-output, camera-normals-output, semantic-output
\end{itemize}

Names in parentheses refer to the internal raw names (can be seen below in the raw graphical visualization too). For reference, the 8 Dronescapes classes are: land, forest, residential, road, little-objects, water, sky and hill. COCO has a total of 133 classes, while Mapillary has a total of 65 classes. All the derived level-2 modalities are done through median on top of either the converted maps or the original maps. Below we provide the conversions that were used to generate the binary maps.

\begin{itemize}
    \item \textbf{Camera normals from depth} We apply a SVD window-based algorithm on top of the marigold depth map, as described in \cite{hartley2003multiple}.
    \item \textbf{vegetation}
    \subitem Mapillary: ["Mountain", "Sand", "Snow", "Terrain", "Vegetation"]
    \subitem COCO: ["tree-merged", "grass-merged", "dirt-merged", "flower", "potted plant", "river", "sea", "water-other", "mountain-merged", "rock-merged"]
    \item \textbf{Sky-and-water}
    \subitem Mapillary: ["Sky", "Water"]
    \subitem Coco: ["sky-other-merged", "water-other", "sea", "river"]
    \item \textbf{Containing}
    \subitem Mapillary: ["Terrain", "Sand", "Mountain", "Road", "Sidewalk", "Pedestrian Area", "Rail Track", "Parking", "Service Lane", "Bridge", "Water", "Curb", "Fence", "Wall", "Guard Rail", "Barrier", "Curb Cut", "Snow"]
    \subitem COCO: ["floor-wood", "floor-other-merged", "pavement-merged", "mountain-merged", "platform", "sand", "road", "sea", "river", "railroad", "grass-merged", "snow", "stairs", "tent"]
    \item \textbf{Transportation}
    \subitem Mapillary: ["Bike Lane", "Crosswalk - Plain", "Curb Cut", "Parking", "Rail Track", "Road", "Service Lane", "Sidewalk", "Bridge", "Tunnel", "Bicyclist", "Motorcyclist", "Other Rider", "Lane Marking - Crosswalk", "Lane Marking - General", "Traffic Light", "Traffic Sign (Back)", "Traffic Sign (Front)", "Bicycle", "Boat", "Bus", "Car", "Caravan", "Motorcycle", "On Rails", "Other Vehicle", "Trailer", "Truck", "Wheeled Slow", "Car Mount", "Ego Vehicle"]
    \subitem COCO: ["bicycle", "car", "motorcycle", "airplane", "bus", "train", "truck", "boat", "road", "railroad", "pavement-merged"]
    \item \textbf{Buildings}
    \subitem Mapillary: ["Building", "Utility Pole", "Pole", "Fence", "Wall"]
    \subitem COCO: ["building-other-merged", "house", "roof"]
    \item \textbf{Building (nearby)} Same as above, but also adds a threshold on top of Mapillary (unscaled depth in 0:1 range) of smaller than 0.3.
    \item \textbf{Safe-landing (geometric)} Uses Camera Normals (SVD) and Marigold depth map and checks the angles on the 3 channels. The logic is: $(v2 > 0.8) * ((v1 + v3) < 1.2) * (depth <= 0.9)$, where v1, v2, v3 are the 3 angles of the camera normals map.
    \item \textbf{Safe-landing (semantic)} Same as geometric, but also includes this mapping.
    \subitem Mapillary: ["Terrain", "Sand", "Snow", "Road", "Lane Marking - General", "Sidewalk", "Bridge", "Pothole", "Catch Basin", "Tunnel", "Parking", "Service Lane", "Pedestrian Area", "Lane Marking - Crosswalk", "On Rails", "Bike Lane", "Crosswalk - Plain", "Mountain", "Vegetation"]
    \subitem COCO: ["grass-merged", "dirt-merged", "sand", "gravel", "flower", "playingfield", "snow", "road", "platform", "railroad", "pavement-merged","mountain-merged", "roof", "tree-merged", "rock-merged"]
\end{itemize}

The full dependency graph exported from the data-pipeline software using graphviz is:

\begin{figure}[h]
    \centering
    \includegraphics[width=1\linewidth]{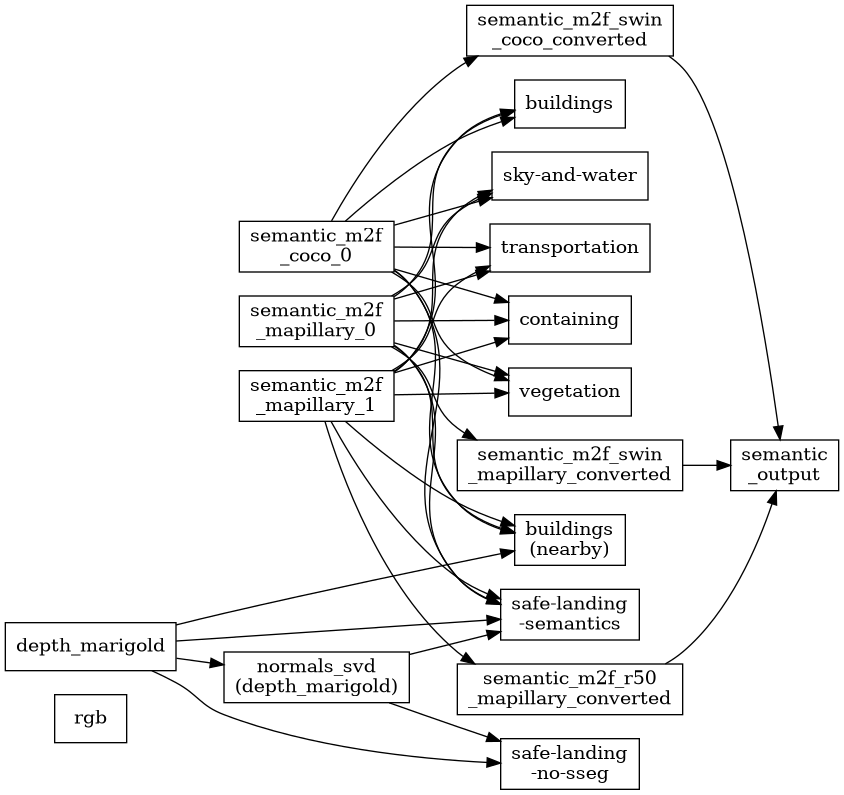}
    \caption{The dependency graph as exported by the data-pipeline tool starting from RGB, through all the pretrained experts and intermediate modalities. For each RGB frame, it results in a sample like \ref{fig:data-pipeline-full}.}
\end{figure}

The Dronescapes2-M+ dataset, as described in Table \ref{tb:dronescapes-variants}, has a total of 80K samples, an increase of 57K samples from the original 23K of Dronescapes-Pseudo1+2. We use the following videos:

\begin{itemize}
    \item{\href{https://huggingface.co/datasets/Meehai/dronescapes2/resolve/main/raw\_data/videos/new\_videos/ovaselu\_DJI\_0372\_540p.mp4?download=true}{ovaselu\_DJI\_0372\_540p.mp4}}
    \item{\href{https://huggingface.co/datasets/Meehai/dronescapes2/resolve/main/raw\_data/videos/new\_videos/politehnica\_DJI\_0741\_a2\_540p.mp4?download=true}{politehnica\_DJ\_0741\_a2\_540p.mp4}}
    \item{\href{https://huggingface.co/datasets/Meehai/dronescapes2/resolve/main/raw\_data/videos/new\_videos/raciu\_DJI\_0418\_540p.mp4?download=true}{raciu\_DJI\_0418\_540p.mp4}}
    \item{\href{https://huggingface.co/datasets/Meehai/dronescapes2/resolve/main/raw\_data/videos/new\_videos/norway\_DJI\_0708\_540p.mp4}{norway\_DJI\_0708\_540p.mp4}}
    \item{\href{https://huggingface.co/datasets/Meehai/dronescapes2/resolve/main/raw\_data/videos/new\_videos/paris\_youtube\_1\_540p.mp4?download=true}{paris\_youtube\_1\_540p.mp4}}
    \item{\href{https://huggingface.co/datasets/Meehai/dronescapes2/resolve/main/raw\_data/videos/new\_videos/sanfrancisco\_youtube\_1\_540p.mp4?download=true}{sanfrancisco\_youtube\_1\_540p.mp4}}
    \item{\href{https://huggingface.co/datasets/Meehai/dronescapes2/resolve/main/raw\_data/videos/new\_videos/riodejaneiro\_youtube\_1\_540p.mp4?download=true}{riodejaneiro\_youtube\_1\_540p.mp4}}
    \item{\href{https://huggingface.co/datasets/Meehai/dronescapes2/resolve/main/raw\_data/videos/new\_videos/sheerness\_youtube\_1\_540p.mp4?download=true}{sheerness\_youtube\_1\_540p.mp4}}
\end{itemize}

For the Dronescapes3 dataset, for which we only extract pseudo-labels for training semantic segmentation distillation models getting to a total of 148K samples, we use the following videos:

\begin{itemize}
    \item{\href{https://huggingface.co/datasets/Meehai/dronescapes2/resolve/main/raw\_data/videos/new\_videos2/voronet\_DJI\_20230918122921\_0102\_D\_540p.mp4?download=true}{voronet\_...\_540p.mp4}}
    \item{\href{https://huggingface.co/datasets/Meehai/dronescapes2/resolve/main/raw\_data/videos/new\_videos2/slanic\_DJI\_20240604124727\_0023\_D\_540p.mp4?download=true}{slanic\_DJI\_20240604124727\_0023\_D\_540p.mp4}}
    \item{\href{https://huggingface.co/datasets/Meehai/dronescapes2/resolve/main/raw\_data/videos/new\_videos2/pietrosa\_DJI\_20240603132329\_0021\_D\_540p.mp4?download=true}{pietrosa\_...\_540p.mp4}}
    \item{\href{https://huggingface.co/datasets/Meehai/dronescapes2/resolve/main/raw\_data/videos/new\_videos2/olanesti\_DJI\_20240924125710\_0005\_D\_540p.mp4?download=true}{olanesti\_...\_540p.mp4}}
    \item{\href{https://huggingface.co/datasets/Meehai/dronescapes2/resolve/main/raw\_data/videos/new\_videos2/horezeu\_DJI\_20240924171535\_0059\_D\_540p.mp4?download=true}{horezeu\_...\_540p.mp4}}
    \item{\href{https://huggingface.co/datasets/Meehai/dronescapes2/resolve/main/raw\_data/videos/new\_videos2/herculane\_DJI\_20240807121925\_0049\_D\_540p.mp4?download=true}{herculane\_...\_540p.mp4}}
    \item{\href{https://huggingface.co/datasets/Meehai/dronescapes2/resolve/main/raw\_data/videos/new\_videos2/comana\_DJI\_20240721134505\_0029\_D\_540p.mp4?download=true}{comana\_...\_540p.mp4}}
    \item{\href{https://huggingface.co/datasets/Meehai/dronescapes2/resolve/main/raw\_data/videos/new\_videos2/castelulcorvinilor\_DJI\_20230915122046\_0040\_D\_540p.mp4?download=true}{castelulcorvinilor\_...\_540p.mp4}}
\end{itemize}

Names shortened above due to spacing issues. The links should work properly.

\subsection{Integration of Expert Generated Modalities with Initial Ground-truth Data}

In order to integrate and align the output modalities with the Dronescapes-Pseudo1 and Dronescapes-Pseudo2 ground truth (SfM and human-annotated semantics) we do the following procedure: We copied the most similar experts, namely we used the median of Mask2Former experts as a proxy for the semantic output modalities (human-annotated), Marigold as a proxy for the metric depth via SfM and the derived camera normals using SVD as a proxy for the camera normals via SfM in the Dronescapes dataset. In order to maintain compatibility, we renormalized the regression tasks' statistics by first subtracting the mean and dividing by the standard deviation (global for all the videos) and then applying the reverse operation by multiplying by the standard deviation of the SfM tasks and adding their mean.

\subsection{Evaluation Metrics}

The Dronescapes evaluation metrics (as described in Section \ref{sec:experiments}) are copied from the original Dronescapes work \cite{marcu2023self}, following the same scripts. For depth estimation, the metric is simply the average L2 * 100, where $L2 = (GT - Prediction)^{2}$.

For semantic segmentation, the metric is defined as the \textit{global} weighted mean intersection over union. Evaluation scripts were officially released alongside the dataset at \url{https://huggingface.co/datasets/Meehai/dronescapes}. A few relevant key points are presented here to ensure correct implementation. The per-class weights are:

\begin{itemize}
    \item land - 0.28172092
    \item forest - 0.30589653
    \item residential - 0.13341699
    \item road - 0.05937348
    \item little-objects - 0.00474491
    \item water - 0.05987466
    \item sky - 0.08660721
    \item hill - 0.06836531
\end{itemize}

These weights were computed as the distribution of pixels on the train set of the original Dronescapes dataset. To get exactly the same results, one needs to follow the same class weights. Then, the Dronescapes-Test dataset contains 3 scenes: barsana, comana and norway. Each of the three scenes are processed independently, so we get three scores: one for each scene. Finally, for each of the three test scenes we do the following process: we accumulate \textit{all} the predictions (i.e. we compute type-1 and type-2 errors) for the entire scene at once. This can be done by computing the stats for each frame, followed by summing them together. Thus, for a single scene, we should have the following 8x4 matrix (after summing the per-frame stats):

\begin{lstlisting}
    class,tp,fp,tn,fn
    land ,XX,XX,XX,XX
    ...
    hill ,YY,YY,YY,YY
\end{lstlisting}

Then compute the intersection over union (or F1 score) using these type-1 and type-2 stats, resulting in 8 independent scores (one per semantic class). Next, we do a weighted sum using the class weights described above resulting in a single number (the per-scene metric). Finally, we do a simple average of all the three scene scores without any scene-level weights to get the final predictions presented throughout this work.

\subsection{Number of frames for each sub-dataset}

Our work provides an extension of the Dronescapes dataset. However, both these datasets are composed of various underlying scenes. Here, we provide the list of the videos in both Dronescapes2 and Dronescapes3 with a count on the number of frames of each such scene, including the ones in the original dataset.

\begin{table}
\scalebox{0.85}{
\begin{tabular}{l|l|l}
\textbf{Scene} & \textbf{Dataset} & \textbf{\#frames} \\
\hline
atanasie\_DJI\_065.. & Dronescapes-Semisup1 & 4501 \\
gradistei\_DJI\_07.. & Dronescapes-Semisup1 & 605 \\
herculane\_DJI\_00.. & Dronescapes-Semisup1 & 363 \\
jupiter\_DJI\_0703.. & Dronescapes-Semisup1 & 726 \\
olanesti\_DJI\_041.. & Dronescapes-Semisup1 & 605 \\
petrova\_DJI\_0525.. & Dronescapes-Semisup1 & 363 \\
slanic\_DJI\_0956\_.. & Dronescapes-Semisup1 & 4500 \\
\hline
atanasie\_DJI\_065.. & Dronescapes-Semisup2 & 4500 \\
gradistei\_DJI\_07.. & Dronescapes-Semisup2 & 484 \\
herculane\_DJI\_00.. & Dronescapes-Semisup2 & 363 \\
jupiter\_DJI\_0703.. & Dronescapes-Semisup2 & 605 \\
olanesti\_DJI\_041.. & Dronescapes-Semisup2 & 484 \\
petrova\_DJI\_0525.. & Dronescapes-Semisup2 & 363 \\
slanic\_DJI\_0956\_.. & Dronescapes-Semisup2 & 4500 \\
\hline
gradistei\_DJI\_07.. & Dronescapes-Semisup1 (val) & 121 \\
herculane\_DJI\_00.. & Dronescapes-Semisup1 (val) & 121 \\
jupiter\_DJI\_0703.. & Dronescapes-Semisup1 (val) & 121 \\
olanesti\_DJI\_041.. & Dronescapes-Semisup1 (val) & 121 \\
petrova\_DJI\_0525.. & Dronescapes-Semisup1 (val) & 121 \\
\hline
barsana\_DJI\_0500.. & Dronescapes-Test & 1452 \\
comana\_DJI\_0881\_.. & Dronescapes-Test** & 1210 \\
norway\_210821\_DJ.. & Dronescapes-Test** & 2941 \\
\hline
barsana\_DJI\_0500.. & Dronescapes-Test* & 36 \\
comana\_DJI\_0881\_.. & Dronescapes-Test* & 30 \\
norway\_210821\_DJ.. & Dronescapes-Test* & 50 \\
\hline
norway\_DJI\_0708\_.. & Dronescapes2 & 4763 \\
ovaselu\_DJI\_0372.. & Dronescapes2 & 9022 \\
paris\_youtube\_1\_.. & Dronescapes2 & 8455 \\
politehnica\_DJI\_.. & Dronescapes2 & 9021 \\
raciu\_DJI\_0418\_5.. & Dronescapes2 & 9022 \\
riodejaneiro\_you.. & Dronescapes2 & 6264 \\
sanfrancisco\_you.. & Dronescapes2 & 5635 \\
sheerness\_youtub.. & Dronescapes2 & 4950 \\
\hline
castelulcorvinil.. & Dronescapes3 & 3121 \\
comana\_DJI\_20240.. & Dronescapes3 & 9315 \\
herculane\_DJI\_20.. & Dronescapes3 & 8204 \\
horezeu\_DJI\_2024.. & Dronescapes3 & 8992 \\
olanesti\_DJI\_202.. & Dronescapes3 & 9765 \\
pietrosa\_DJI\_202.. & Dronescapes3 & 9762 \\
slanic\_DJI\_20240.. & Dronescapes3 & 8985 \\
voronet\_DJI\_2023.. & Dronescapes3 & 9766 \\
\end{tabular}
} 
\caption{The detailed number of frames for each sub-dataset. \\
* The subset of human-annotated frames. Semantic segmentation evaluation is run only on these. \\
** No SfM reconstructions on these scenes, thus no evaluation is done for depth or camera normals.}
\end{table}

\bibliographystyle{elsarticle-num}
\bibliography{references}
\end{document}